\documentclass{article}

\usepackage[final]{corl_2026}

\usepackage{amsmath,amssymb,amsfonts}
\usepackage{bm}
\usepackage{graphicx}
\usepackage{booktabs}
\usepackage{multirow}
\usepackage{tabularx}
\usepackage{xspace}
\usepackage{xcolor}
\usepackage{algorithm}
\usepackage{algorithmic}
\usepackage{tikz}
\usepackage{pgfplots}
\usepackage{listings}
\usepackage{caption}
\usepackage{subcaption}
\usepackage{url}
\usepackage{siunitx}
\usepackage{mathtools}

\graphicspath{{figs/}{supps_fig/}}
\pgfplotsset{compat=1.18}

\newcommand{\eg}{\emph{e.g.}\xspace}
\newcommand{\ms}{\mathcal{S}}
\newcommand{\mt}{\mathcal{T}}

\newcommand{\ours}{UniLatent\xspace}

\newcommand{\yufei}[1]{}
\newcommand{\haoran}[1]{}
\newcommand{\daniel}[1]{}
\newcommand{\weiduo}[1]{}
\newcommand{\yue}[1]{}
\newcommand{\yidu}[1]{}
\newcommand{\todo}[1]{}

\title{\ours: Cross-Embodiment Transfer\\via Latent Observation Alignment}

\author{
  Yufei Ding$^{1*}$\quad Yi Du$^{1*}$\quad Weiduo Yuan$^{2}$\quad Hao Jiang$^{2}$ \\
  \textbf{Dylan Goetting$^{3}$\quad Lawrence Yunliang Chen$^{3}$\quad Daniel Seita$^{2\dagger}$\quad Yue Wang$^{2\dagger}$} \\[2pt]
  $^{1}$Stanford University\qquad $^{2}$University of Southern California\qquad $^{3}$UC Berkeley \\
  {\small $^{*}$Equal contribution\qquad $^{\dagger}$Equal advising}
}
\hypersetup{pdfauthor={Yufei Ding, Yi Du, Weiduo Yuan, Hao Jiang, Dylan Goetting, Lawrence Yunliang Chen, Daniel Seita, Yue Wang}}

\begin{document}
\maketitle

\begin{abstract}
Robot datasets are costly to collect and are usually tied to a specific
embodiment, making it difficult to reuse demonstrations across robots with
different geometry, appearance, and kinematics. We introduce \ours, a cross-embodiment transfer framework that reduces the embodiment-induced observation gap in a shared latent space.
\ours renders paired source--target robot observations in simulation and aligns a source visual encoder to a frozen target encoder. A policy trained on encoded source demonstrations can then run directly on target observations, with no pixel-level image translation. 
Across three simulation tasks and two target robot arms, UniLatent reaches up to 93\% zero-shot success, exceeding the strongest pixel-translation baseline by over 30 points on average. In real-world xArm-to-Franka transfer, fine-tuning with only 10 target demonstrations per task yields 27/30 successful rollouts. Code, videos and data are available at
\url{https://selina2023.github.io/unilatent/}.

\end{abstract}

\begin{figure}[H]
  \centering
  \includegraphics[width=\linewidth]{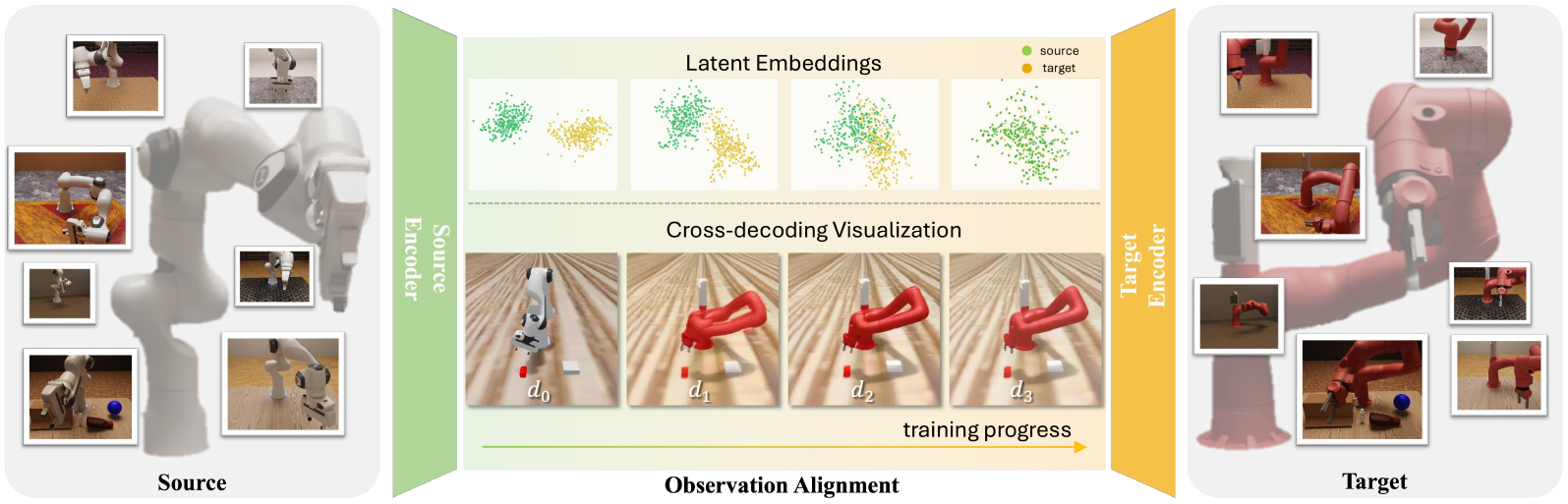}
  \caption{\textbf{UniLatent aligns robot observations for cross-embodiment transfer.}
UniLatent renders paired source--target images and aligns a source encoder to a target encoder, forming a shared latent space that reduces embodiment-induced visual gaps. \textit{Top:} source and target embeddings increasingly overlap during training. \textit{Bottom:} cross-decoded source observations become more target-consistent as alignment improves. 
}
  \label{fig:teaser}
\end{figure}

\section{Introduction}
Large-scale and diverse data is a key driver of recent progress in robot learning~\cite{FMs_robotics_2023_survey_1,FMs_robotics_2023_survey_2,zhong2025surveyVLA}, and recent efforts have produced sizable manipulation datasets~\cite{open_x_embodiment_rt_x_2023,khazatsky2024droid,bu2025agibot_iros}. However, existing robot data is highly imbalanced across embodiments, which concentrates on a small set of popular arms and grippers, while many other platforms remain underrepresented. Meanwhile, collecting real-world data is often costly and embodiment-specific, since demonstrations largely depend on the robots' morphology, end-effector geometry, reachable workspace, appearance, and kinematics. As robot hardware continues to diversify and advance, recollecting the same task for every new embodiment does not scale. This motivates methods that reuse source-robot demonstrations and allow policies to transfer to unseen robots with little or no target data.

A key challenge in cross-embodiment transfer is the embodiment-induced visual gap. In many existing datasets with third-person camera views, the robot often occupies a large portion of the image, and changing the arm or gripper alters not only appearance but also self-occlusion, contact geometry, and the visibility of task-relevant objects, which causes noticeable performance degradation when deploying a vision-based policy on a new robot~\citep{chen2024mirage,chen2024roviaug}. Prior work addresses this gap by editing source demonstration frames through masking~\cite{lepert2024shadow}, inpainting~\cite{chen2024mirage}, or diffusion-based generation~\cite{chen2024roviaug} to match the target visual domain. While intuitive, pixel-space translation can introduce errors in segmentation, robot geometry, and contact regions, creating systematic discrepancies from true target observations. Moreover, translated frames can create a synthetic visual domain that differs
from real target observations, causing policies to overfit to translation artifacts and limiting their ability to generalize across tasks and viewpoints.

We introduce UniLatent, which reduces the embodiment-induced observation gap in latent space rather than pixel space (Fig.~\ref{fig:teaser}). In simulation, we render large-scale paired observations in which source and target robots reach matched end-effector poses under the same camera and
randomized scenes. These pairs train a source encoder to map each source image into the latent space of a frozen, pretrained target encoder, so that matched poses on the two robots receive the same representation. A policy 
trained on encoded source demonstrations is then deployed on the target robot by encoding its observations with the same frozen target encoder, with no image synthesis at test time.

Our contributions are two-fold. \textbf{(i) Paired rendering and observation alignment.} We introduce a simulation-based pipeline that generates large-scale paired observations of different robots at matched end-effector poses and uses them to align visual encoders across embodiments. \textbf{(ii) Latent-space policy learning.} We train flow-matching policies on source demonstrations encoded in the aligned latent 
space, enabling zero-shot transfer to unseen target embodiments in simulation and data-efficient few-shot transfer in the real world. In simulation, \ours raises mean target success from 44.5\% with the strongest pixel-translation baseline to 76.0\%; in the real world, it completes 27/30 rollouts versus 23/30 for RoVi-Aug pretraining and 11/30 for target-only training.

\section{Related Work}
\textbf{Visual adaptation in robotics.} Vision-based policies are brittle under visual shifts caused by lighting, texture, backgrounds, and viewpoints. Domain randomization broadens the training distribution by varying rendering factors in simulation~\cite{tobin2017domainrandomization}, while sim-to-real methods learn mappings between visual domains~\cite{james2019sim,ho2021retinagan,rao2020rl}. Recent data augmentation methods further use photometric, semantic, or generative edits on real robot data to improve robustness~\cite{bharadhwaj2024roboagent,yu2023scaling,tian2024vista,chen2024semanticallycontrollable,yuan2025roboengineplugandplayrobotdata}. These approaches are effective for appearance and viewpoint changes, but most assume that the robot embodiment is fixed. In our setting, the robot itself changes, so the visual shift is tied to kinematics, gripper shape, self-occlusion, and contact cues.

\textbf{Policy learning across embodiments.} Robot-generalist policies train across many robots or condition on embodiment descriptors~\cite{black2025pi0,octomodelteam2024,intelligence2025pi05,chen2018hardwareconditioned}. Other work studies shared action abstractions, universal actions, or large multi-embodiment datasets~\cite{yang2024pushinglimits,doshi2024scalingcrossembodied,salhotra2023actionmismatch,zheng2025universalactions}. These methods are promising when broad real-robot coverage is available, but such coverage is expensive and uneven across platforms. \ours targets a complementary regime: given demonstrations from a source robot and a desired target robot, we learn a target-consistent observation space from simulation-rendered pairs and reuse the source data without requiring a large multi-robot dataset.

\textbf{Cross-embodiment transfer.} A related line studies cross-embodiment imitation from third-person or human video prompts, including skill discovery from human and robot videos~\cite{xu2023xskillcrossembodimentskill}, imitation under mismatched execution~\cite{kedia2025oneshotimitationmismatchedexecution}, and video-conditioned policy learning~\cite{jain2024vid2robotendtoendvideoconditionedpolicy}. Human-to-robot and human-to-humanoid transfer bridges an even larger embodiment gap, using egocentric human video~\cite{kareer2025egomimic,kareer2025emergencehumanrobottransfer}, human-video editing~\cite{lepert2025phantom}, whole-body shadowing~\cite{fu2024humanplus}, or sim-to-real RL from a human demonstration~\cite{lum2025crossinghumanrobotembodimentgap}. These works are complementary to our setting: they interpret human or video demonstrations as task specifications or retarget human motion, whereas we study robot-to-robot reuse of source demonstrations for a specified target embodiment.

More recent methods reduce embodiment shift by modifying images. Mirage cross-paints robot pixels~\cite{chen2024mirage}; RoVi-Aug synthesizes robot and viewpoint augmentations~\cite{chen2024roviaug}; and Shadow masks robot regions to avoid robot-specific appearance~\cite{lepert2024shadow}. While these methods can improve over direct transfer, pixel editing can discard gripper or contact information or introduce geometry that is inconsistent with the real target observation. \ours instead aligns encoders and acts in latent space. This removes the need for test-time image translation and lets the target observation encoder define the representation used by the deployed policy. Cross-embodiment and domain-invariant representation learning has also explored alignment objectives in different settings, \eg, inverse RL~\cite{zakka2021xirl}, time-contrastive video representations~\cite{sermanet2018tcn}, and adversarial sim-to-real alignment~\cite{tanwani2020dirl}. Closest to ours, Wang et al.~\cite{wang2024crossembodiment} align an adapted-domain latent space to a frozen anchor domain without pixel translation, using unpaired adversarial and cycle-consistency objectives. \ours instead constructs explicit correspondence by rendering both robots at matched TCP poses, which reduces alignment to supervised feature regression onto a frozen pretrained target encoder.

\begin{figure}[t]
    \centering
    \includegraphics[width=\linewidth]{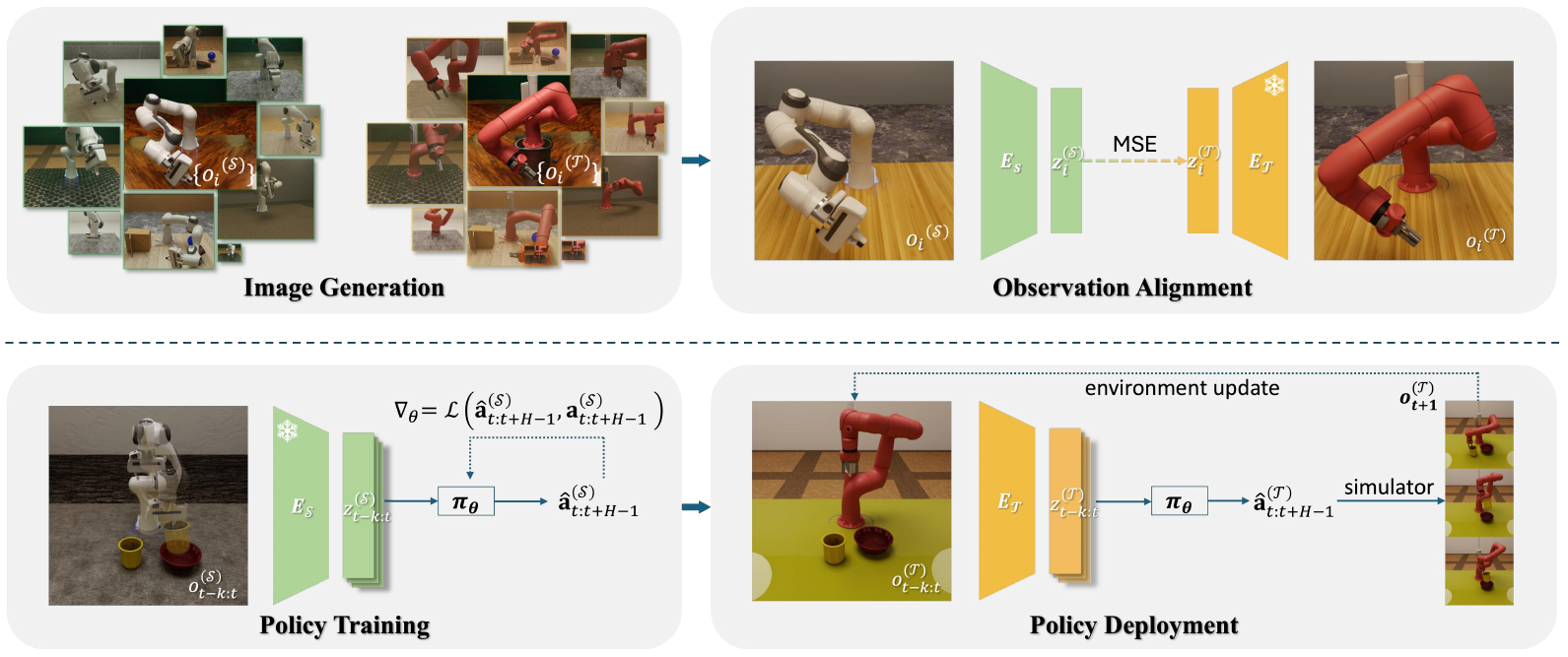}
    \caption{\textbf{Pipeline overview of \ours.} We render paired (source, target) observations $\{(o_i^{(\ms)}, o_i^{(\mt)})\}$ for matched end-effector poses and use them to align encoders by training $E_\ms$ to match a fixed $E_\mt$ in a shared latent space. A latent policy $\pi_\theta$ is then trained on source demonstrations and deployed on the target robot by encoding target observations with $E_\mt$ and calling $\pi_\theta$.}
    \label{fig:pipeline}
    \vspace{-9pt}
\end{figure}

\section{Problem Statement}
\label{sec:prob_stmt}
We study robot-to-robot cross-embodiment imitation, where demonstrations
collected on a source robot $R_{\mathcal S}$ are used to learn a policy for a
different target robot $R_{\mathcal T}$ with minimal or no target-robot
demonstrations. Each robot observes the task through RGB images
$o^{(R)} \in \mathcal{O}^{(R)}$ and executes actions
$a^{(R)} \in \mathcal{A}^{(R)}$ through a shared Cartesian end-effector
interface. In our setting, the robots are single-arm manipulators with
parallel-jaw grippers, but they may differ in geometry, kinematics, gripper
appearance, self-occlusion, and robot-object contact cues.
We assume access to $M$ source demonstrations
$\{\tau_i\}_{i=1}^M$, where each trajectory
$\tau_i=\{(o_{i,t}^{(\mathcal S)},a_{i,t}^{(\mathcal S)})\}_{t=1}^{T_i}$
is collected on $R_{\mathcal S}$ under a fixed third-person camera. The goal is
to learn a target policy
$\pi_{\mathcal T}:\mathcal{O}^{(\mathcal T)}\rightarrow
\mathcal{A}^{(\mathcal T)}$ that solves the same task on $R_{\mathcal T}$. We further assume access to kinematic and visual models of both robots
(\eg, URDF or USD assets) in a simulator with inverse kinematics. These models provide paired observations for visual alignment, without
requiring target demonstrations or task-success supervision on the target robot. Source demonstrations need not be collected for alignment; alignment consumes only robot models, inverse kinematics, and rendering.

\section{Method: \ours}
\label{sec:method}
Given source demonstrations $\{\tau_i\}_{i=1}^{M}$ collected on $R_\ms$, \ours{} learns a latent-space policy that transfers to $R_\mt$ with few or no target demonstrations.
The pipeline has four stages (Fig.~\ref{fig:pipeline}). We first (i) render paired source--target observations in simulation (Sec.~\ref{rendering}) and (ii) align a source encoder to a frozen target encoder (Sec.~\ref{alignment}). We then (iii) train a latent policy on encoded source demonstrations and (iv) deploy it on $R_\mt$ through the target encoder (Sec.~\ref{policy}).

\subsection{Paired Image Generation}
\label{rendering}
\begin{figure}[t]
    \centering
    \includegraphics[width=1.00\linewidth]{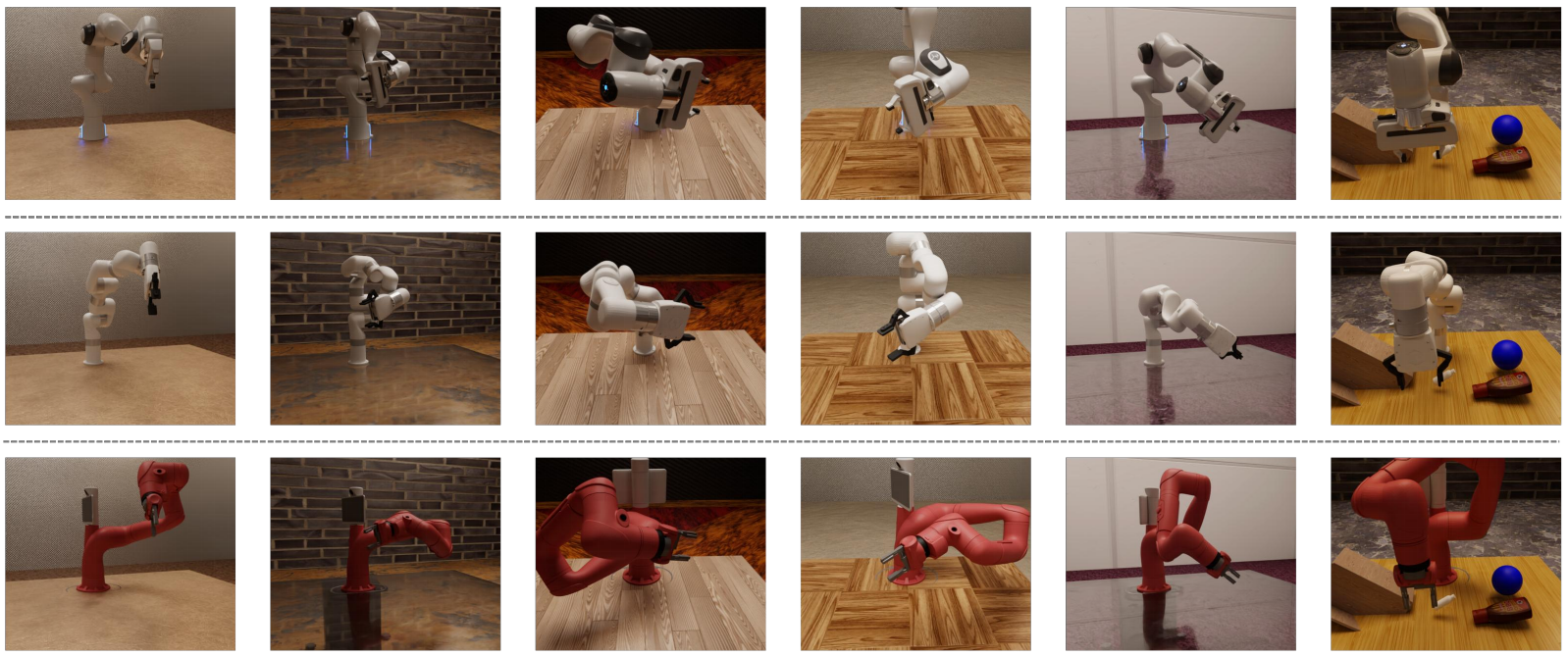}
    \caption{\textbf{Paired image generation.} Examples of simulation-rendered paired observations for three robots (top to bottom): Franka, xArm, and Sawyer. For each sampled end-effector pose, we render the robots under randomized backgrounds, lighting, and scene variations while preserving action-level correspondence between paired frames.}
    \label{fig:paired_img_gen}
    \vspace{-7pt}
\end{figure}

We sample manipulation-relevant end-effector poses $\{p_i\}$ within the robots' reachable workspace, solve inverse kinematics for both embodiments, and render both robots from the same camera under the same randomized scene. The source and target images in each pair share the same task-relevant pose but differ in robot geometry and appearance. This produces paired RGB images $\{(o_i^{(\ms)},o_i^{(\mt)})\}_{i=1}^{N}$ in IsaacLab/RoboVerse~\cite{mittal2025isaaclab,geng2025roboverse} (Fig.~\ref{fig:paired_img_gen}).

The pose generator combines broad workspace sampling with a manipulation prior so that rendered pairs emphasize poses that can occur in demonstrations and avoid unreachable or visually uninformative configurations. We match poses at the tool-center point (TCP), defined as the midpoint between
the parallel-jaw fingertips, apply robot-specific TCP offsets, and retain only
samples for which inverse kinematics succeeds for both embodiments. Domain randomization varies backgrounds, lighting, object assets, distractors, and small camera perturbations while preserving pairwise action correspondence. These choices make the supervision robust to appearance changes without requiring the renderer to synthesize photorealistic target demonstrations. Our \emph{generic} pairs (about 100k) use this broad randomized sampling, with roughly 20\% of poses drawn from ManiSkill, LIBERO, RLBench, and RoboSuite trajectories that share no tasks with our evaluation suite. Optional \emph{task-specific} pairs (about 10k) additionally use the evaluation task's object assets and poses.

\subsection{Observation Alignment}
\label{alignment}
We instantiate a source encoder $E_\ms$ and target encoder $E_\mt$ from DINOv2-base~\cite{oquab2024dinov2learningrobustvisual}. The target encoder is frozen and defines the deployment representation. For each paired rendering, we train only $E_\ms$ to match target features:
\begin{equation}
\mathcal{L}_{\mathrm{align}}=\|E_\ms(o_i^{(\ms)})-E_\mt(o_i^{(\mt)})\|_2^2.
\end{equation}
In practice the loss is applied to the encoder tokens used by the policy, including global and patch-level features. Freezing $E_\mt$ anchors the latent space to the representation used at deployment and prevents target-side drift: target images are encoded by exactly the model that defines the alignment target. Training only $E_\ms$ asks each source image to produce the feature of its paired target image, so the policy sees consistent representations when trained on source data and deployed on the target robot.

\subsection{Latent Policy Training and Deployment}
\label{policy}
After alignment, each source demonstration is encoded as $z_t^{(\ms)}=E_\ms(o_t^{(\ms)})$. A policy $\pi_\theta$ conditions on a history $\mathbf{z}_{t-k+1:t}^{(\ms)}$ and predicts an end-effector action chunk $\mathbf{a}_{t:t+H-1}^{(\ms)}$. Actions are represented in a shared Cartesian end-effector space with position, orientation, and a binary gripper command. 

We use a conditional flow-matching model~\cite{liu2022rectifiedflow,lipman2023flowmatchinggenerativemodeling}: starting from $\mathbf{a}_0\sim\mathcal{N}(0,I)$ and a demonstrated chunk $\mathbf{a}_1$, it learns the velocity field
$\mathcal{L}_{\mathrm{FM}}=\mathbb{E}\big[\|v_\theta(\mathbf{a}_u,u\mid\mathbf{z}_{t-k+1:t}^{(\ms)})-v^\star(\mathbf{a}_u,u)\|_2^2\big].$
\label{ssec:policy_deployment}
At inference, target observations are encoded as $z_t^{(\mt)}=E_\mt(o_t^{(\mt)})$. The learned flow is integrated with 10 solver steps, and the resulting pose command is executed by the target robot controller.

\section{Experiments}
\label{sec:exps}
We design the experiments to answer three questions. \textbf{(Q1)} How large is the embodiment gap when only the action interface is shared? \textbf{(Q2)} Is explicit alignment necessary, or does a single pretrained encoder shared by both robots suffice? \textbf{(Q3)} Under the same alignment, is it better to learn the policy in latent space or on translated pixels? Sec.~\ref{subsec:exp-sim} addresses all three questions with zero-shot transfer in simulation, using 100 rollouts per setting. Sec.~\ref{subsec:real-world-exps} evaluates few-shot real-world transfer from xArm to Franka, including a comparison with the original diffusion-based RoVi-Aug pipeline.

\subsection{Simulation-Based Experiments}
\label{subsec:exp-sim}

\begin{figure}[H]
    \centering
    \setlength{\unitlength}{1pt}
    \begin{picture}(242,152)(0,0)
        \put(37,0){\includegraphics[height=152pt,trim=70bp 0bp 0bp 0bp,clip]{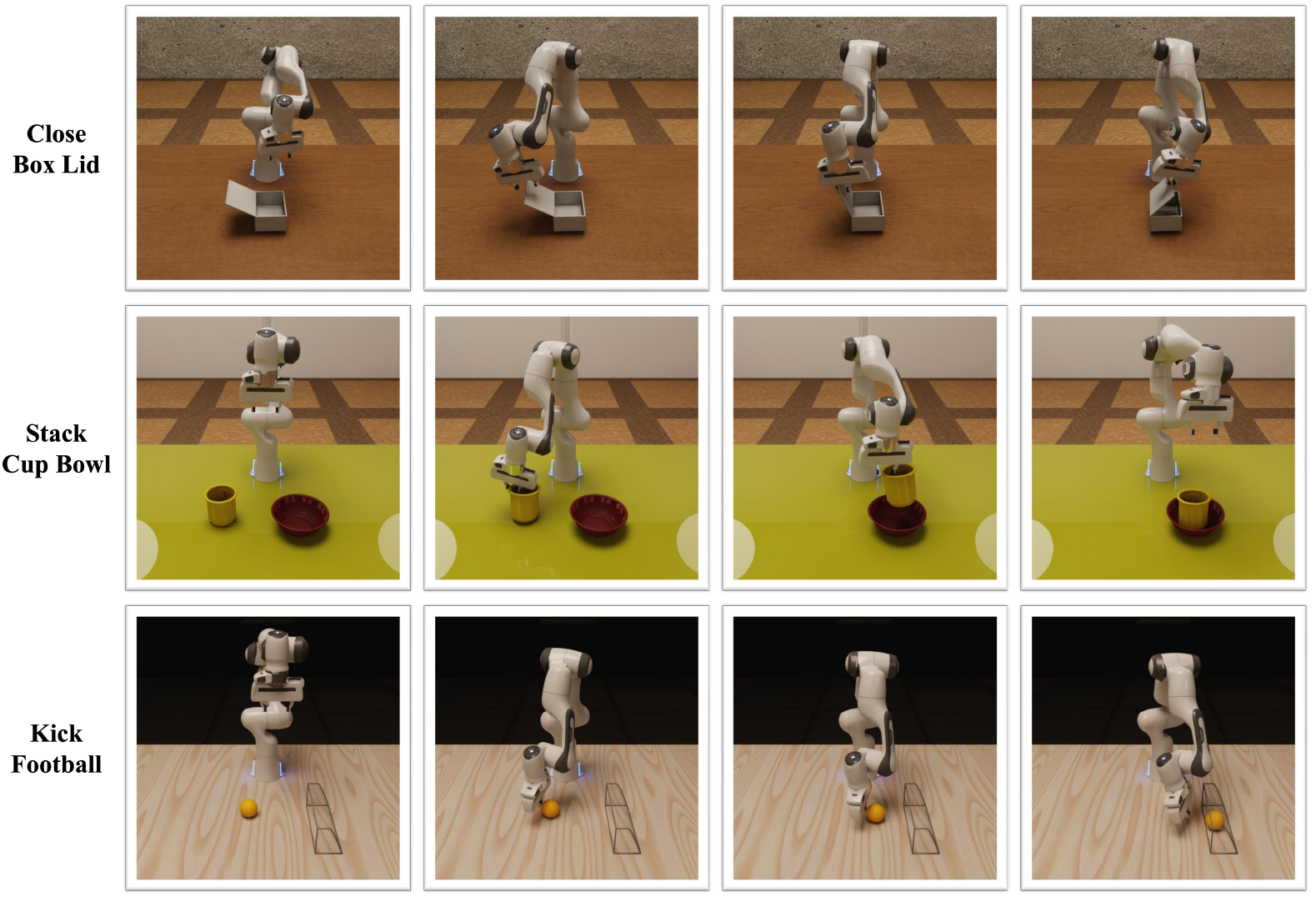}}
        \put(0,124){\makebox(38,0){\small\bfseries\begin{tabular}[c]{@{}c@{}}Close\\Box Lid\end{tabular}}}
        \put(0,76){\makebox(38,0){\small\bfseries\begin{tabular}[c]{@{}c@{}}Stack\\Cup into\\Bowl\end{tabular}}}
        \put(0,22){\makebox(38,0){\small\bfseries\begin{tabular}[c]{@{}c@{}}Kick\\Football\end{tabular}}}
    \end{picture}
    \caption{\textbf{Simulation tasks.} Example frames from our simulation experiments. Each row shows representative frames of a Franka demonstration under a fixed third-person camera.}
    \label{fig:sim_tasks}
    \vspace{-5pt}
\end{figure}

\textbf{Robots.} We use Franka Panda as the source robot and evaluate zero-shot transfer to xArm7 and Sawyer. All three are 7-DoF arms with parallel-jaw grippers, but they differ in link geometry, joint limits, gripper mechanism, and appearance. All robots share a Cartesian TCP-pose action interface with a binary gripper command, mapped to each robot through robot-specific TCP offsets and gripper joint targets (Appendix~\ref{supp:robots}). Source and target share the task, camera, and non-robot scene factors, so the robot embodiment is the only changed factor. Existing datasets such as Open X-Embodiment~\cite{open_x_embodiment_rt_x_2023} and DROID~\cite{khazatsky2024droid} cannot support this setting, as each demonstration is recorded on a single robot.

\textbf{Tasks.} We design three simulated tasks in the RoboVerse framework~\cite{geng2025roboverse} with an Isaac Sim backend (Fig.~\ref{fig:sim_tasks}). The tasks are designed to test transfer under different manipulation modes: articulated contact, grasp-and-place precision, and nonprehensile pushing. Success is determined from the final object state. \textit{Close Box Lid}: a box starts with its lid open at $135^\circ$, and the robot closes the lid about its hinge; success requires a final hinge angle below $14^\circ$. \textit{Stack Cup into Bowl}: the robot picks up a small cup and places it into a bowl; success requires all predefined cup keypoints to lie inside a bounding box within the bowl. \textit{Kick Football}: the robot pushes a ball (4\,cm radius) into a goal; success requires the ball center to lie inside a bounding box in the goal mouth. Object initial poses are randomized within task-specific ranges for both training and evaluation.

\textbf{Data and evaluation.} For each task, we collect 300 Franka demonstrations in simulation by teleoperating the simulated robot through a DexCap-based interface~\cite{wang2024dexcapscalableportablemocap}. We replay each demonstration in the simulator with physics enabled, render ray-traced RGB observations from a fixed third-person camera, and discard replays that fail the success checker. Each method is evaluated with 100 closed-loop rollouts per task and robot, using initial states drawn from random seeds disjoint from training; a rollout fails if the success condition is not met within 500 control steps. For alignment, we render about 100k generic pairs plus 10k task-specific pairs (Sec.~\ref{rendering}). Policies use a frozen aligned encoder, a conditional UNet flow-matching head, history $k=3$, action horizon $H=8$, and 66\,Hz control (Appendix~\ref{sec:supp-impl}).

\textbf{Baselines.} We compare against three baselines, all trained on the same 300 Franka demonstrations with the same action space and evaluated with the same protocol. (i) \textit{Direct Policy Transfer} \textbf{(Q1)} trains a standard visuomotor policy, with an MLP visual encoder learned end-to-end on source demonstrations, and deploys it on the target robot without visual adaptation. (ii) \textit{DINOv2 w/o alignment} \textbf{(Q2)} encodes both source and target observations with one shared frozen DINOv2 encoder and trains the same latent policy as \ours, with the same pooling, but without paired-rendering alignment. It tests whether generic pretrained features are already invariant to the robot embodiment. (iii) \textit{RoVi-Aug*} \textbf{(Q3)} follows the RoVi-Aug pipeline~\cite{chen2024roviaug}: source demonstrations are translated into the target robot's visual domain, and a pixel-space policy trained on the translated images is deployed on target RGB observations. RoVi-Aug* changes only the translator. Whereas RoVi-Aug synthesizes the target robot with diffusion-based cross-painting, RoVi-Aug* encodes each source image with our aligned source encoder $E_\ms$ and decodes the features with the pretrained Representation Autoencoder decoder~\cite{zheng2025diffusiontransformersrepresentationautoencoders}, which maps DINOv2 features back to RGB. Because $E_\ms$ is aligned to the frozen target encoder, the decoded image shows the target robot. We adopt this translator for its translation quality: on our robot pairs, diffusion-based generation distorted robot links and joint structure under challenging poses, and segmentation masking (Shadow~\cite{lepert2024shadow}) removed gripper--object contact regions, whereas cross-decoding produced the most faithful target-robot images (Appendix~\ref{sec:supp-translation-qualitative}). RoVi-Aug* thus gives pixel-space transfer a strong translator while sharing \ours's paired renderings, aligned encoder, demonstrations, and action supervision, so the main remaining difference is whether the policy learns from translated pixels or from aligned latent features. We evaluate the original diffusion-based RoVi-Aug in the real-world experiments (Sec.~\ref{subsec:real-world-exps}).

\begin{table}[!htbp]
\centering
\small
\setlength{\tabcolsep}{3.6pt}
\renewcommand{\arraystretch}{1.05}
\resizebox{\linewidth}{!}{%
\begin{tabular}{lccc ccc ccc}
\toprule
\multirow{2}{*}{\textbf{Task \& Setup}}
& \multicolumn{3}{c}{\textbf{Close Box Lid}}
& \multicolumn{3}{c}{\textbf{Stack Cup into Bowl}}
& \multicolumn{3}{c}{\textbf{Kick Football}} \\
\cmidrule(lr){2-4}\cmidrule(lr){5-7}\cmidrule(lr){8-10}
& Franka & xArm & Sawyer & Franka & xArm & Sawyer & Franka & xArm & Sawyer\\
\midrule
Direct Policy Transfer & 97 & 2 & 5 & 80 & 0 & 0 & 90 & 1 & 1\\
DINOv2 w/o align. & 97 & 40 & 16 & 89 & 3 & 0 & 93 & 11 & 1\\
RoVi-Aug* & 97 & 93 & 54 & 89 & 15 & 5 & 93 & 60 & 40\\
\textbf{\ours{} (ours)} & \textbf{97} & \textbf{93} & \textbf{81} & \textbf{89} & \textbf{65} & \textbf{51} & \textbf{93} & \textbf{89} & \textbf{77}\\
\bottomrule
\end{tabular}%
}
\caption{\textbf{Zero-shot transfer results in simulation.} Policies are trained on Franka demonstrations and evaluated on Franka (in-domain) and zero-shot on xArm and Sawyer. We compare direct policy transfer, a shared frozen DINOv2 encoder without alignment, RoVi-Aug* (our adaptation of RoVi-Aug), and the full \ours{} pipeline. Entries report success rates (\%) over 100 rollouts.}

\label{tab:sim_exp}
\end{table}

\textbf{Results.} In Table~\ref{tab:sim_exp}, the Franka column reports in-domain source performance as a reference, and the xArm and Sawyer columns report zero-shot transfer without target demonstrations. For Q1, direct transfer achieves only 0--5\% target success despite 80--97\% in-domain success, showing that matching DoF and gripper type does not imply a small visual embodiment gap. For Q2, frozen DINOv2 features improve some xArm settings but remain insufficient, reaching at most 16\% on Sawyer, so explicit alignment is necessary. For Q3, averaged over the six target task--robot settings, RoVi-Aug* achieves 44.5\% success, while \ours achieves 76.0\%. \ours matches RoVi-Aug* on xArm \textit{Close Box Lid} and improves the other five settings. The largest gains occur on \textit{Stack Cup into Bowl}: from 15\% to 65\% on xArm and from 5\% to 51\% on Sawyer.

\begin{figure}
    \centering
    \includegraphics[width=0.52\linewidth]{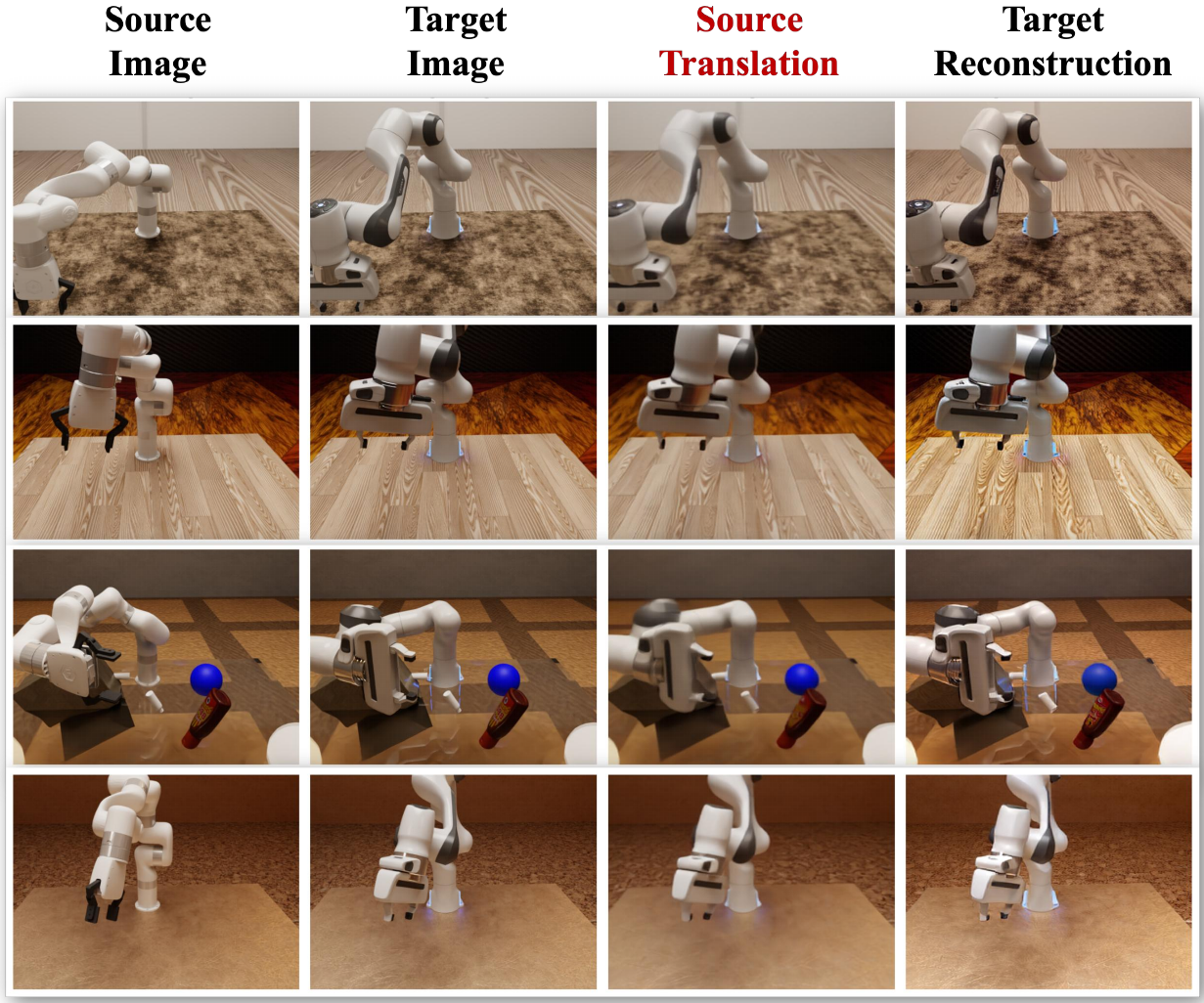}
    \caption{\textbf{Pixel-space translation examples.} Each row shows source observation $o^{(S)}$, its target pair $o^{(T)}$, decoded source translation $D_{\mathcal{T}}(E_{\mathcal{S}}(o^{(S)}))$ used by RoVi-Aug*, and target reconstruction $D_{\mathcal{T}}(E_{\mathcal{T}}(o^{(T)}))$.}
    \label{fig:decoding}
    \vspace{-5pt}
\end{figure}

\textbf{Latent vs.\ pixel space.} Cross-decoding preserves global layout but can introduce errors around the wrist, gripper opening, and manipulated objects (Fig.~\ref{fig:decoding}). RoVi-Aug* thus trains on decoded images but executes on real target images, whereas \ours trains and executes on features from the same frozen target encoder, avoiding this train--test mismatch.

\subsection{Real-World Experiments}
\label{subsec:real-world-exps}

\begin{figure}[H]
    \centering
    \begin{minipage}[t]{0.55\linewidth}
        \centering
        \vspace{0pt}
        \includegraphics[width=\linewidth]{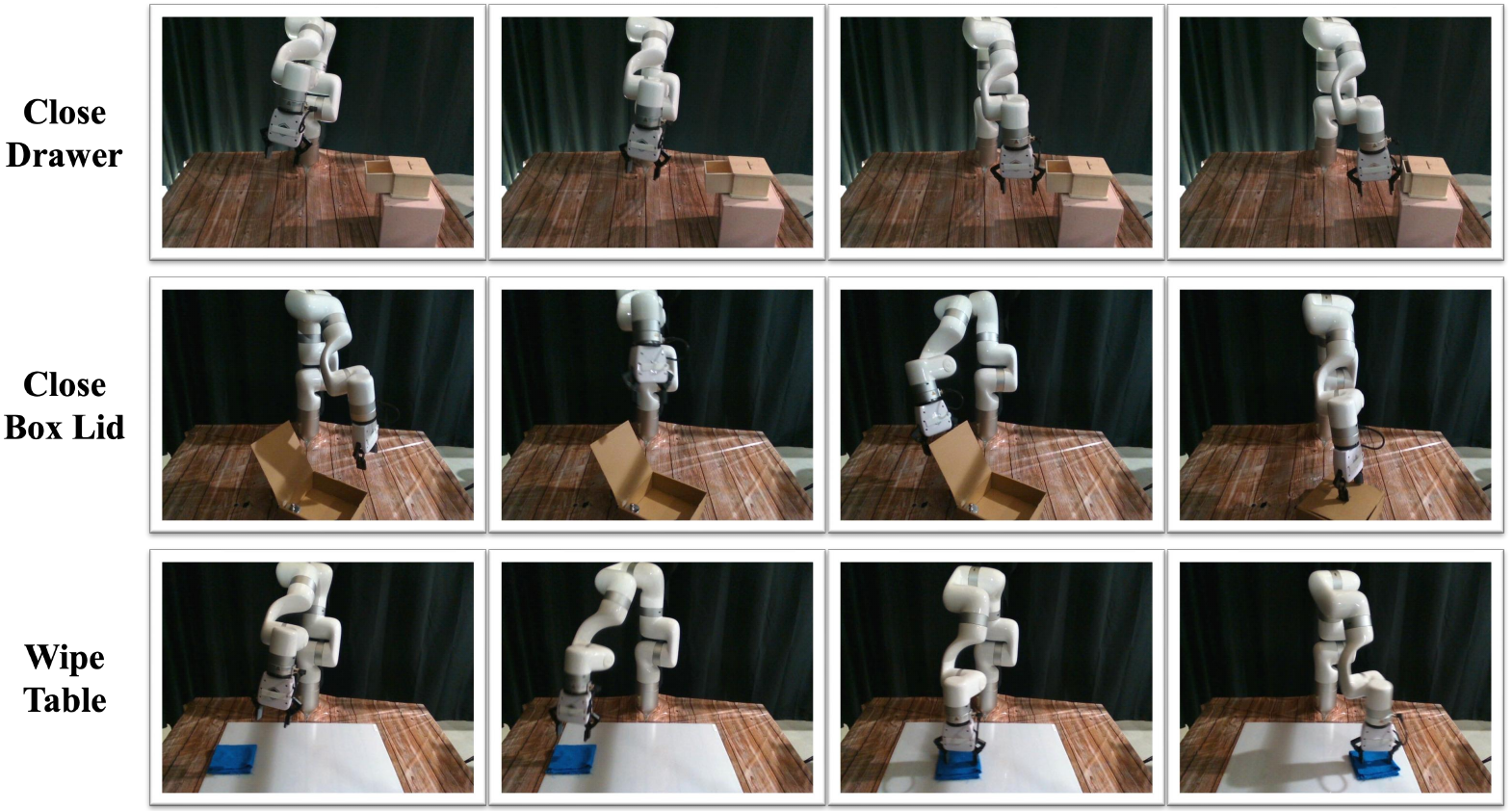}
        \caption{\textbf{Real-world tasks.} Each row shows representative frames from a teleoperated demonstration collected on xArm under a fixed third-person camera.}
        \label{fig:real_tasks}
    \end{minipage}\hfill
    \begin{minipage}[t]{0.43\linewidth}
        \centering
        \vspace{0pt}
        \footnotesize
        \setlength{\tabcolsep}{2pt}
        \renewcommand{\arraystretch}{1.06}
        \begin{tabularx}{\linewidth}{@{}>{\raggedright\arraybackslash}Xccc@{}}
            \toprule
            \textbf{Setting} & \textbf{Drawer} & \textbf{Box} & \textbf{Wipe} \\
            \midrule
            Source-only (xArm) & 10/10 & 8/10 & 10/10 \\
            Target-only (few-shot, Franka) & 7/10 & 2/10 & 2/10 \\
            RoVi-Aug pretrain + few-shot & 9/10 & 9/10 & 5/10 \\
            \textbf{Cross-emb pretrain + few-shot (ours)} & \textbf{10/10} & \textbf{9/10} & \textbf{8/10} \\
            \bottomrule
        \end{tabularx}
        \captionof{table}{\textbf{Few-shot real-world transfer.} \textit{Source-only} is an in-domain xArm reference. Rows 2--4 are evaluated on Franka and use the same $M_\mt=10$ Franka demonstrations. \textit{Cross-emb pretrain + few-shot (ours)} is \ours.}
        \label{tab:real_finetune}
    \end{minipage}
    \vspace{-4pt}
\end{figure}
\textbf{Setup.} We transfer from a real xArm source robot to a Franka target robot on \textit{Close Drawer} (push a drawer closed), \textit{Close Box Lid} (push an open lid down), and \textit{Wipe Table} (wipe the table with a cloth from right to left, following the RoVi-Aug setting~\cite{chen2024roviaug}); see Fig.~\ref{fig:real_tasks}. We collect 150 teleoperated xArm demonstrations for \textit{Close Drawer} and 61 for each of the other two tasks at 10\,Hz, with a RealSense D435i third-person camera at $640\times480$ whose extrinsics are identical for the two robots, so that, as in simulation, the embodiment is the main changed factor. A single task-agnostic encoder, aligned using only about 100k generic paired renderings, is shared by all real-world tasks. The target encoder is off-the-shelf DINOv2, pretrained on real images and kept frozen, so simulation supervision only trains $E_\ms$ toward a fixed real-image representation. Since alignment is learned purely in simulation and the physical robots also differ in controllers and calibration, we evaluate few-shot transfer, fine-tuning with $M_\mt=10$ Franka demonstrations per task. Each setting is evaluated with 10 rollouts per task from randomized initial object configurations, and a rollout succeeds if the task criterion is met within 25\,s.

\textbf{Compared settings.} \textit{Source-only} trains and evaluates on xArm as an in-domain upper reference. The remaining three settings are evaluated on Franka and use the same 10 target demonstrations per task; they differ only in how source data is used. \textit{Target-only} trains from scratch. \textit{RoVi-Aug pretrain + few-shot} follows the original RoVi-Aug pipeline~\cite{chen2024roviaug}, providing a direct comparison with the published method. Each xArm frame is segmented with Grounded-SAM2~\cite{ren2024grounded,ravi2024sam2segmentimages}, and the xArm is replaced by a Franka at the same end-effector pose, synthesized by a ControlNet-conditioned diffusion model~\cite{ControlNet_2023_ICCV} that we train on rendered xArm--Franka pairs, since RoVi-Aug releases no checkpoints for our robot pair (Appendix~\ref{sec:supp-real-roviaug-preprocess}). A pixel-space policy is pretrained on the translated xArm demonstrations and then fine-tuned on the Franka data. \textit{Cross-emb pretrain + few-shot} pretrains our latent policy on aligned xArm observations and fine-tunes with the same Franka data.

\textbf{Results.} Table~\ref{tab:real_finetune} reports 27/30 successful target rollouts for \ours, compared with 23/30 for RoVi-Aug pretraining and 11/30 for target-only training. RoVi-Aug pretraining improves over target-only training on all three tasks, confirming that source data is useful when bridged to the target domain. Relative to RoVi-Aug, \ours improves \textit{Close Drawer} from 9/10 to 10/10 and \textit{Wipe Table} from 5/10 to 8/10, while matching 9/10 on \textit{Close Box Lid}. A single encoder aligned only on generic simulated pairs thus provides a useful initialization for all three real-world tasks at the evaluated 10-demonstration budget. Rollout videos, including failures, are on the project website.

\subsection{Multi-Source, Viewpoint, and Background Transfer}
\label{subsec:additional}
We examine three extensions on \textit{Close Box Lid} (Table~\ref{tab:transfer_analyses}).

\vspace{-3pt}
\begin{center}
\small
\setlength{\tabcolsep}{0pt}
\renewcommand{\arraystretch}{1.02}

\begin{tabular}{@{}p{0.58\linewidth}@{\hspace{0.035\linewidth}}p{0.385\linewidth}@{}}
\begin{minipage}[t]{\linewidth}
\vspace{0pt}
\centering
\textbf{(a) Multi-source transfer to xArm}

\begin{tabular*}{\linewidth}{@{\extracolsep{\fill}}lcc@{}}
\toprule
\textbf{Source robot(s)} & \textbf{\# Demos} & \textbf{Success (\%)} \\
\midrule
Sawyer          & 100 & 64 \\
Franka          & 100 & 63 \\
Sawyer          & 200 & 84 \\
Franka          & 200 & \textbf{92} \\
Sawyer + Franka & 200 & 91 \\
\bottomrule
\end{tabular*}
\end{minipage}
&
\begin{minipage}[t]{\linewidth}
\vspace{0pt}
\centering
\textbf{(b) Viewpoint shift}

\begin{tabular*}{\linewidth}{@{\extracolsep{\fill}}lcc@{}}
\toprule
\textbf{Alignment} & \multicolumn{2}{c}{\textbf{Deploy view}} \\
\cmidrule(lr){2-3}
 & \textbf{v1} & \textbf{v2} \\
\midrule
Matched view & \textbf{97} & 62 \\
Cross-view   & 96 & \textbf{89} \\
\bottomrule
\end{tabular*}
\end{minipage}
\end{tabular}

\vspace{3pt}
\begin{minipage}{0.72\linewidth}
\centering
\textbf{(c) Simulated background generalization}

\begin{tabular*}{\linewidth}{@{\extracolsep{\fill}}lccc@{}}
\toprule
\textbf{Training / evaluation backgrounds} & \textbf{Franka (src.)} & \textbf{xArm} & \textbf{Sawyer} \\
\midrule
Matched & \textbf{97} & \textbf{93} & \textbf{81} \\
Randomized source / five unseen scenes & 88.6 & 85.6 & 80.0 \\
\bottomrule
\end{tabular*}
\end{minipage}

\vspace{-2.8pt}
\captionof{table}{\textbf{Additional transfer analyses on \textit{Close Box Lid} (success, \%).}
(a) Multi-source transfer to xArm from Sawyer, Franka, or both source robots; demo counts denote total source demonstrations, and all sources align to the same frozen target encoder $E_\mt$.
(b) Cross-view paired rendering improves deployment on the shifted view v2.
(c) Success under matched backgrounds and under source-background randomization, averaged over five held-out simulated scenes.}
\label{tab:transfer_analyses}
\vspace{-4.8pt}
\end{center}
\textbf{Multi-source transfer.} We align separate Franka and Sawyer source encoders to the same frozen xArm target encoder and train one policy on their combined demonstrations. Using 100 demonstrations from each source achieves 91\% success, compared with 63--64\% for either 100-demonstration source alone, and is comparable to Franka-only training with 200 demonstrations (92\%). Heterogeneous sources can thus be combined without changing deployment.

\textbf{Viewpoint-shift transfer.} We render paired observations from two camera viewpoints (v1, v2) of the \emph{same} robot and align v1 features to v2. A matched-view policy achieves 97\% on v1 but only 62\% on v2. Cross-view alignment achieves 96\% and 89\%, showing that paired rendering also accommodates systematic viewpoint shifts.

\textbf{Background generalization.} We render Franka source demonstrations with randomized backgrounds and evaluate the transferred policies in five held-out simulated scenes. Mean success is 88.6\% on Franka, 85.6\% on xArm, and 80.0\% on Sawyer, compared with 97\%, 93\%, and 81\% under matched backgrounds. These results suggest that visual diversity added on the source side carries over to the target policy through the aligned latent space. This experiment covers simulated background shifts; real-world clutter and lighting changes remain untested.

\subsection{Analysis}

\begin{figure}[H]
    \centering
    \includegraphics[width=1.0\linewidth]{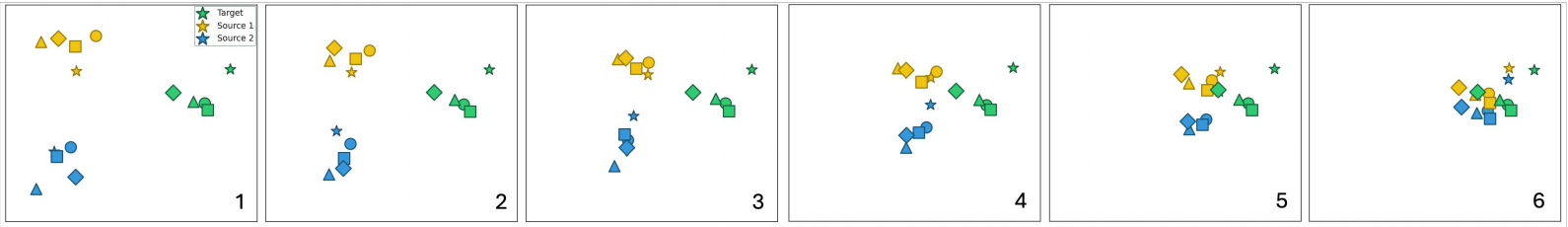}
    \caption{\textbf{t-SNE of latent embeddings during alignment.} Each point denotes a latent embedding of a rendered image; colors indicate robot embodiments, and marker shapes indicate paired rendered samples. Across training (1$\rightarrow$6), embeddings from different robots become more aligned in the latent space.}
    \label{fig:tsne}
    \vspace{-10pt}
\end{figure}

Fig.~\ref{fig:tsne} shows embodiment-separated features from Franka, Sawyer, and xArm becoming increasingly overlapping during alignment. Paired-pose coverage also matters: with the regression objective unchanged, manipulation-relevant pose sampling yields 93\%/81\% success on xArm/Sawyer, versus 60\%/32\% with unconstrained sampling (Table~\ref{tab:pose-prior-close-box}, Appendix~\ref{sec:supp-additional}). Further analyses of pooling, diffusion versus flow policies, and pixel-space translation are provided in Appendix~\ref{sec:supp-additional}.

\section{Limitations}
\ours currently targets 7-DoF single-arm robots with parallel-jaw grippers and a shared end-effector action interface. Substantially different morphologies, such as dexterous hands, humanoids, or mobile manipulators, and tasks requiring tool-specific contact would require additional retargeting and contact modeling. Our tasks were designed so that both robots perform the same task in the same scene; applying \ours to external source datasets such as Open X-Embodiment remains to be demonstrated. The method also assumes that the target robot can complete source tasks in a comparable workspace; demonstrations outside the target reachable set may need trajectory modification. Finally, our background-shift evaluation is limited to one simulated task, and our real-world experiments use a single physical setup; robustness to real-world clutter, lighting, distractors, and broader object variation remains unestablished.

\section{Conclusions}
We present \ours, a latent observation alignment framework for cross-embodiment robot learning. By aligning source observations to a frozen target representation with paired simulation renderings, \ours deploys source-trained policies on target robots without test-time image translation. Results in simulation and the real world indicate that latent alignment is an effective and data-efficient path to cross-embodiment reuse among arms with similar morphologies. We release code, the rendering pipeline, datasets, and encoder checkpoints to support reproducibility.

\acknowledgments{We thank researchers at UC Berkeley and USC for helpful discussions. Daniel Seita acknowledges generous support from Samsung Research America and NSF grant \#2447397.}

\clearpage
\bibliography{references}

\clearpage
\appendix

\section{Appendix Overview}
\label{sec:supp-overview}

This appendix provides implementation details, experimental protocols, baseline specifications, and additional analyses for \ours. It is organized into four parts: (i) implementation details of the \ours~pipeline, (ii) experimental setup details for simulation and real-world evaluations, (iii) baseline implementations and pixel-space translation analysis, and (iv) additional analyses and ablations that examine key design choices in representation alignment and latent policy learning, including a study of generalization to unseen backgrounds and scenes.

\paragraph{Implementation details (\S\ref{sec:supp-impl}).}
We describe the full \ours~pipeline in detail. We first explain how paired source-target robot observations are generated in simulation, including manipulation-relevant pose sampling, tool-center-point matching, inverse-kinematics filtering, domain randomization, and camera configuration. We then describe target-anchored observation alignment with DINOv2 encoders, including the encoder architecture, feature normalization, alignment objective, and training recipe. Finally, we provide the latent-space policy implementation, including the end-effector action representation, flow-matching policy parameterization, training procedure, deployment on the target robot, and compute cost.

\paragraph{Experiment setup details (\S\ref{sec:supp-exp}).}
We specify the simulation and real-world evaluation settings used in the main experiments. For simulation, we provide the tabletop environment setup, task definitions, object randomization ranges, state-based success checkers, rollout protocol, robot control parameters, teleoperation interface, and replay-and-filtering pipeline for demonstration collection. For real-world experiments, we describe the source and target robot setups, camera configuration, data-collection procedure, few-shot target demonstration protocol, task definitions, evaluation randomization, success criteria, and rollout termination rules.

\paragraph{Baselines and pixel-space translation analysis (\S\ref{sec:supp-baselines}).}
We provide implementation details for the baselines used to evaluate cross-embodiment transfer, including direct policy transfer, a frozen DINOv2 policy without observation alignment, and RoVi-Aug$^\ast$ as a pixel-space translation baseline. We also compare representative pixel-space translation strategies, including diffusion-based robot-to-robot generation, segmentation-based masking, and cross-decoding of aligned features through a pretrained RAE decoder. This analysis characterizes how pixel-level methods preserve or distort robot geometry, gripper state, scene context, and contact regions, and motivates the use of latent-space policy execution in \ours.

\paragraph{Additional ablation studies and explorations (\S\ref{sec:supp-additional}).}
We analyze several design choices in \ours and study its generalization beyond the default deployment setting. First, we study pooling and feature design for DINOv2 representations, comparing fixed pooling rules and learnable attention pooling under frozen and unfrozen backbone settings. Second, we compare diffusion and flow-matching policies in the aligned latent space under matched architectures and training protocols. Third, we ablate the pose generator used for paired rendering by comparing manipulation-relevant pose sampling with unconstrained pose sampling. Finally, we evaluate transfer under background and scene shifts between source training and target deployment, showing that source-side background randomization is inherited by the transferred target policy through the shared latent space.

\section{\ours~Implementation Details} \label{sec:supp-impl}
\subsection{Paired Image Generation}

\paragraph{Pose sampling.}\label{sec:supp-impl-pose-sampling}

To generate paired rendered observations for alignment training, we first need a distribution over \textit{plausible} gripper poses that commonly arise in table-top manipulation. We define poses at the tool center point (TCP) instead of the robot's terminal link, since the TCP is the manipulation-relevant point that should be matched across robots.

For our parallel-jaw grippers, we define the TCP as a frame rigidly attached to the gripper whose origin is the midpoint between the two jaw tips\footnote{Strictly speaking, the TCP is a coordinate frame, not just a point. In our setup, its origin is the midpoint between the two jaw tips, and its orientation is rigidly inherited from the gripper. Thus, the midpoint specifies the TCP origin, while the full TCP pose is determined once the gripper orientation is given.}. This choice accounts for the fixed geometric offset between the terminal link frame used by cuRobo's robot model and the point in space where interaction with objects is centered. It is this manipulation-relevant point, not the terminal link itself, that should be aligned across robots for paired image generation. 

We derive a plausible pose distribution by referring to diverse table-top manipulation tasks in ManiSkill~\citep{mu2021maniskillgeneralizablemanipulationskill}, LIBERO~\citep{liu2023liberobenchmarkingknowledgetransfer}, RLBench~\citep{james2019rlbenchrobotlearningbenchmark}, and RoboSuite~\citep{zhu2025robosuitemodularsimulationframework}. Concretely, we parameterize each sampled TCP pose by its Cartesian position $(x, y, z)$ and its orientation, represented using a tilt angle $\theta$ measured from pointing straight down (along the $-z$ axis) together with an azimuthal angle $\phi$ around the vertical axis. As a simplification, we assume an independent distribution of each dimension of the pose. Although this assumption is not necessarily true, we found that in practice it yields plausible and diverse poses.

We implement two sampling methods. The first method samples from a parametric distribution derived from analyzing demonstration data. The TCP position coordinates follow independent Gaussian distributions: $x \sim \mathcal{N}(0.70, 0.20^2)$, $y \sim \mathcal{N}(0.0, 0.23^2)$, and $z \sim \mathcal{N}(0.35, 0.12^2)$, with soft clamping to stay within physical workspace limits. The tilt angle $\theta$ follows a truncated exponential distribution $P(\theta) \propto \exp(-\theta/\lambda)$ with decay constant $\lambda = 35^\circ$ and maximum angle $120^\circ$, which strongly favors gripper orientations pointing mostly downward as commonly observed in manipulation tasks. The azimuthal angle $\phi$ is sampled uniformly from $[0, 2\pi)$. The sampled TCP pose is then converted to joint angles via inverse kinematics. The gripper state is independently randomized, with 70\% probability of being open and 30\% closed.

The second method samples directly from the empirical distribution of end-effector poses extracted from demonstrations across diverse tasks in existing datasets. We load all demonstration trajectory files, extract joint poses at regular intervals, compute the corresponding TCP poses via forward kinematics, and then solve IK for the target robot. This approach preserves the natural correlation structure between pose dimensions present in real manipulation trajectories, at the cost of being limited to the support of the demonstration data.

We combine the two sampling methods at a roughly 8:2 proportion, with the majority coming from the parametric distribution and the remainder from the empirical demonstration distribution. Importantly, the tasks we use for downstream experiments are entirely self-collected and do not appear in any of the existing benchmarks (ManiSkill, LIBERO, RLBench, or RoboSuite) from which we derived our pose distribution. This separation ensures that our alignment training allows the robot to learn from reasonable, manipulation-relevant poses without overfitting to any specific task or demonstration dataset, preserving the generalization capability of the learned representation.

\paragraph{Inverse kinematics.}
To ensure all generated poses are kinematically feasible, we employ cuRobo~\citep{sundaralingam2023curoboparallelizedcollisionfreeminimumjerk} as our inverse kinematics solver. Given a sampled TCP pose (position and orientation), cuRobo solves for joint angles that achieve this end-effector configuration while respecting the robot's joint limits. We initialize the IK solver with seed configurations derived from each robot's default joint positions, which biases solutions toward natural manipulation-ready arm configurations rather than awkward or self-colliding poses. The IK solver processes poses in batches for computational efficiency. Poses for which IK fails to find a valid solution are rejected from the dataset. When retargeting to multiple target robot embodiments, we apply an intersection filter: only frames where IK succeeds for all target robots are retained. This filtering ensures that each retained pose is kinematically reachable by all robot embodiments, so that corresponding configurations can be rendered across robots.

In practice, we observe that xArm has a relatively compact kinematic structure and thus a smaller reachable workspace compared to Franka and Sawyer. Poses with large height values that are feasible on the latter two robots often fail to be retargeted to xArm. We accommodate this limitation in our task design: the tasks we evaluate on (Close Box Lid, Stack Cup into Bowl, Kick Football) involve relatively small objects and operate in a lower workspace region, ensuring fairness across all robot embodiments.

For Sawyer, the IK seed configuration is particularly critical. The Sawyer's arm is long, slender, and highly articulated, meaning a single end-effector pose can correspond to multiple valid joint configurations. However, some geometrically valid IK solutions place the elbow or forearm in configurations that obstruct the gripper's intended motion. During paired image generation this is not immediately problematic since we do not require task completion, but during inference, a poorly-seeded initial pose can leave the arm blocking subsequent movements, rendering the task unsolvable. To ensure consistency between training and inference, we initialize the IK solver with each robot's default joint positions as a reference seed. This biases solutions toward manipulation-ready configurations where the arm is positioned out of the way, ensuring that the joint configurations shown in training images match what the robot will produce at inference time.

\paragraph{Pose matching details.}\label{sec:supp-impl-pose-matching}
To ensure action equivalence across embodiments, we match poses at TCP, defined as the point between the fingertips, instead of at the end-effector link (the wrist). Since different robots have different gripper lengths, matching wrist poses directly would result in fingertips at different positions in space, which is problematic for manipulation where contact points matter. Instead, we apply a fixed transformation to convert the source robot's wrist pose to its TCP pose. We then solve IK for the target robot such that its TCP reaches the same $\mathrm{SE}(3)$ pose.

\paragraph{Domain randomization.}
To reduce overfitting to simulator-specific visual appearances and better match real-world visual variability, we apply extensive domain randomization during paired image generation using the RoboVerse simulation infrastructure.

For scene objects, we augment existing RoboVerse assets with 3D object models from the YCB dataset, converted to USD format for Isaac Sim.

We randomize surface materials for all scene elements: floors are sampled from carpet, wood, stone, concrete, and architectural textures; walls from architectural panels, wall board, masonry, paint, and composite materials; and table surfaces from wood, stone, plastic, ceramic, and metal finishes. Lighting conditions are randomized by varying the intensity (within plausible ranges for indoor scenes), color temperature (3000K–6000K to simulate different light sources from warm tungsten to cool daylight), and position of a main ceiling disk light and four corner sphere lights.

We additionally apply small perturbations to camera pose ($\pm 5$cm in position, $\pm 2^\circ$ in orientation) and look-at point to simulate natural variation in camera placement. This randomization encourages the learned representation to capture pose correspondence, not scene-specific visual features.

\paragraph{Camera configurations.}
We use a consistent camera setup across all robot embodiments to ensure viewpoint equivalence in paired images. The camera is positioned at a fixed location viewing the tabletop workspace from a front-diagonal perspective, with the look-at point centered on the manipulation area. We render images at $640\times480$ resolution with camera intrinsics calibrated to match a real Intel RealSense camera (focal length 6.06mm, horizontal aperture 6.4mm), providing a wide field of view that captures the robot arm throughout its workspace. Critically, for each pose tuple, all robots are rendered from the identical camera viewpoint—same extrinsics and intrinsics—so that the only visual difference between paired images is the robot morphology itself, not the observation perspective. Small camera perturbations ($\pm 5$cm in position, $\pm 2^\circ$ in orientation) are applied as part of domain randomization to improve robustness, but these perturbations are consistent within each rendered frame, preserving the correspondence between paired observations.

\subsection{Observation Alignment Learning}

\paragraph{Encoder architecture.}
Both $E_\ms$ and $E_\mt$ are instantiated from the DINOv2-base model with register tokens (\texttt{facebook/dinov2-with-registers-base})~\citep{oquab2024dinov2learningrobustvisual}. The backbone is a ViT-B/14 architecture with 12 transformer layers, hidden dimension 768, and patch size $14\times14$. Input images are resized to $224\times224$ via bicubic interpolation and normalized with the ImageNet channel-wise mean $\mu_{\text{ImageNet}}=[0.485, 0.456, 0.406]$ and standard deviation $\sigma_{\text{ImageNet}}=[0.229, 0.224, 0.225]$. We extract features from the last hidden state of the DINOv2 backbone. The output token sequence has length 261: 1 CLS token, 4 register tokens, and $16\times16=256$ patch tokens. We strip the register tokens and use both the CLS token, a single 768-dimensional vector, and the 256 patch tokens, each 768-dimensional, for alignment. The patch token sequence of shape $(B, 256, 768)$ is reshaped into a 2D spatial feature map of shape $(B, 768, 16, 16)$, with $B$ denoting the batch size. We disable the learnable affine parameters (scale and bias) of DINOv2's final LayerNorm, retaining only the non-parametric mean-centering and variance-normalization. Both encoders are initialized from the same pretrained DINOv2 checkpoint; $E_\mt$ is then frozen and $E_\ms$ is the only module updated during alignment training.

\paragraph{Alignment objective.}
We minimize the MSE between corresponding source and target embeddings. Let
$z_{\mathrm{cls}}^{(R)} \in \mathbb{R}^{B \times 768}$ and
$z_{\mathrm{patch}}^{(R)} \in \mathbb{R}^{B \times 768 \times 16 \times 16}$
denote the CLS token and reshaped patch feature map produced by encoder $E_R$, where
$R \in \{S,T\}$. The alignment loss is
\begin{equation}
\mathcal{L}_{\mathrm{align}}
=
\left\| z_{\mathrm{cls}}^{(S)} - z_{\mathrm{cls}}^{(T)} \right\|_2^2
+
\left\| \hat{z}_{\mathrm{patch}}^{(S)} - \hat{z}_{\mathrm{patch}}^{(T)} \right\|_2^2 .
\end{equation}

The pipeline uses two distinct types of normalization. First, input RGB images are normalized with the standard ImageNet pixel statistics expected by the pretrained DINOv2 checkpoint. Second, after feature extraction, we standardize the reshaped patch feature map using feature-channel statistics computed from the frozen DINOv2 encoder over ImageNet images:
\begin{equation}
\hat{z}_{b,c,u,v}
=
\frac{
z_{b,c,u,v} - \mu^{\mathrm{feat}}_{c}
}{
\sqrt{(\sigma^{\mathrm{feat}}_{c})^2 + \epsilon}
},
\qquad
\epsilon = 10^{-5}.
\end{equation}
Here, $c$ indexes a DINOv2 feature channel, not an RGB image channel. Thus, $\mu^{\mathrm{feat}}_{c}$ and $\sigma^{\mathrm{feat}}_{c}$ are feature-level statistics and are distinct from the RGB ImageNet pixel statistics used for input preprocessing. The same feature standardization is applied to both source and target patch features before computing the patch-level MSE.

\paragraph{Training recipe.}
We use the Adam optimizer with learning rate $10^{-4}$ and momentum parameters $\beta = (0.5, 0.9)$. The learning rate follows a cosine annealing schedule that decays to a minimum of $10^{-6}$. We train for 25 epochs with a batch size of 10 on a single A6000 GPU, with validation performed twice per epoch. Training uses float32 precision. During training, we apply camera-quality degradation augmentations to the source images using Albumentations~\cite{buslaev2020albumentations}: at each training sample, one augmentation is randomly selected from Gaussian noise, ISO noise, motion blur, JPEG compression, brightness/contrast perturbation, and hue/saturation shift, with an overall application probability of 0.9. The target image is not augmented, since the target encoder $E_\mt$ is frozen and serves as the fixed reference. This asymmetric augmentation encourages the source encoder to learn features that are robust to realistic imaging variations while preserving a stable target representation.

\paragraph{Time cost and practical considerations.}
Paired rendering and encoder alignment are offline costs. In our experiments, the alignment dataset contains approximately $100$k generic paired renderings, optionally supplemented with $10$k task-specific pairs, and the source encoder is trained for $25$ alignment epochs on a single NVIDIA A6000 GPU. An encoder aligned on generic pairs can be reused across downstream tasks, as in our real-world experiments, where only the task policy is trained from the corresponding source demonstrations. This contrasts with real-world demonstration collection or task-specific synthetic trajectory generation, which must be repeated for each new task. The offline rendering and alignment cost is therefore amortized across tasks and does not affect deployment, where target observations are encoded directly by the frozen target encoder.

\subsection{Latent-Space Policy Training and Deployment}

\paragraph{Action representation.}\label{sec:supp-imp-action-representation}
The policy uses an 8-dimensional action representation in the end-effector space, formulated as $a = [p, q, g]$. Translation is represented as a 3D position vector $p = (p_x, p_y, p_z)$ in the robot base frame. Rotation is encoded as a unit quaternion $q = (q_w, q_x, q_y, q_z)$. Because quaternions $q$ and $-q$ represent the identical spatial rotation, we canonicalize the training dataset to a consistent hemisphere by enforcing $q_w \ge 0$.

The gripper state is represented by a binary scalar $g \in \{0, 1\}$, where 0 indicates a closed gripper and 1 indicates an open gripper. To map continuous demonstration data into this binary space, we compute the L2 distance between the recorded gripper joint state and the robot's canonical fully-open and fully-closed configurations. Because grasped objects physically prevent the gripper from reaching its absolute closed limit, we apply an asymmetric distance threshold that biases the classification toward the closed state. This abstraction enables cross-embodiment transfer by decoupling the action space from robot-specific gripper joint kinematics.

During training, we independently apply min-max normalization to all eight dimensions to scale them to the range $[-1, 1]$. For the $i$-th action dimension $a_i$, the normalized value is computed as $a_{i, \mathrm{norm}} = 2 \cdot \frac{a_i - a_{i, \min}}{a_{i, \max} - a_{i, \min}} - 1$, where $a_{i, \min}$ and $a_{i, \max}$ are the minimum and maximum values observed in the training data for that specific dimension. Standard safeguards are applied for degenerate dimensions where $a_{i, \max} = a_{i, \min}$. During inference, the network outputs are de-normalized back to their original scales. To ensure a valid rotation before execution on the robot, the predicted quaternion is explicitly re-normalized to unit length, such that $q \leftarrow q / \|q\|$.

\paragraph{Flow matching policy.}
The policy $\pi_\theta$ is instantiated as a conditional flow-matching \citep{lipman2023flowmatchinggenerativemodeling, liu2022rectifiedflow} generative model that maps a window of latent observations to an end-effector action chunk. Given a source demonstration $\tau=\{(o_t^{(\ms)}, a_t^{(\ms)})\}_{t=1}^{T}$ on source robot $\ms$ with length $T$, we encode each observation with the aligned source encoder to obtain the latent vector $z_t^{(\ms)} = E_\ms(o_t^{(\ms)})$ and form a latent conditioning window $\mathbf{z}^{(\ms)}_{t-k+1:t}$. In practice, we train on fixed-length action chunks: the supervision target is $\mathbf{a}^{(\ms)}_{t:t+H-1}$, and we define $\mathbf{a}_1\triangleq\mathbf{a}^{(\ms)}_{t:t+H-1}$ with horizon $H$ (we use $k=3$ and $H=8$ in our experiments). Concretely, we sample an initial action chunk $\mathbf{a}_0 \sim \mathcal{N}(0,I)$ and construct an interpolation path $\mathbf{a}_u$ between $\mathbf{a}_0$ and $\mathbf{a}_1$ for $u\in[0,1]$. We use the linear path $\mathbf{a}_u = (1-u)\mathbf{a}_0 + u\mathbf{a}_1$, which induces a constant target velocity $v^\star(\mathbf{a}_u,u)=\mathbf{a}_1-\mathbf{a}_0$. The model is parameterized by a conditional UNet backbone (implemented as a 1D UNet over the action-chunk temporal dimension) and predicts a conditional velocity field $v_\theta(\mathbf{a}_u,u \mid \mathbf{z}^{(\ms)}_{t-k+1:t})$. Training minimizes the standard flow-matching objective
\begin{equation*}
\mathcal{L}_{\mathrm{FM}}=
\mathbb{E}\big[\|v_\theta(\mathbf{a}_u,u \mid \mathbf{z}^{(\ms)}_{t-k+1:t})-v^\star(\mathbf{a}_u,u)\|_2^2\big],
\end{equation*}
where the expectation is taken over $\mathbf{a}_0\sim\mathcal{N}(0,I)$ and $u\sim\mathrm{Uniform}[0,1]$. At inference time, action chunks are generated by integrating the learned flow field from $u=0$ to $u=1$ with 10 solver steps; in our implementation we use a midpoint (second-order Runge--Kutta) ODE solver. The UNet uses channel dimensions $(256,512,1024)$, kernel size $5$, GroupNorm with $8$ groups, and a flow-time embedding dimension of $128$.

\paragraph{Training details.}
We train the flow-matching policy using AdamW optimizer with learning rate $10^{-4}$, $\beta = (0.95, 0.999)$, and weight decay $10^{-6}$. The learning rate follows a cosine schedule with $500$ warmup steps. We use a batch size of $256$ and train for $200$ epochs. For the temporal structure, the policy conditions on a history of the most recent $k=3$ latent observations and predicts action chunks of length $H=8$. During training, the model conditions on encoded latent observations provided as a global conditioning signal. We apply Exponential Moving Average (EMA) with maximum decay $0.9999$ for stable inference. The flow-matching loss is a simple MSE between the predicted and target velocity fields, where all action dimensions (translation, rotation, gripper) are weighted equally. The aligned source encoder $E_{\ms}$ is kept frozen throughout policy training.

\paragraph{Deployment details.}
\textbf{Simulation.} At inference time, the policy operates in a closed-loop manner. Given the current RGB observation $o_t^{\mt}$, the policy maintains a sliding window of the $k=3$ most recent latent observations. When the action cache is exhausted, the policy encodes the observation history through the frozen target encoder $E_{\mt}$, producing latent features that are passed to the flow-matching U-Net. The flow field is integrated from $u=0$ to $u=1$ using the midpoint ODE solver with $10$ integration steps, generating an action chunk of $H=8$ end-effector commands. These $8$ actions are then executed sequentially—one per simulation step—before the policy is queried again for a new chunk. 

Each predicted action consists of an absolute Cartesian pose (3D position + quaternion rotation) and a binary gripper command, which are converted to joint positions via inverse kinematics (cuRobo) before being sent to the robot-specific low-level controller. To maintain temporal consistency within an action chunk, each IK solution is seeded with the previous solution. If IK fails for a particular target pose, the robot holds its current joint configuration for that timestep. Episodes terminate upon task success or after a maximum of 500 timesteps.

\textbf{Real world.} Consistent with simulation setup, we utilize the identical model architecture for real-world inference. The policy conditions on a sliding window of $k=3$ historical observations to predict an action chunk of $H =  8$ timestamps. We deploy the policy at a control frequency of 10 Hz. Episodes are terminated and recorded as a failure if the task is not successfully completed within 250 steps (25 seconds).

\section{Experiment Setup Details}
\label{sec:supp-exp}

\subsection{Tasks and Environments}

\paragraph{Simulation environment.}
Our simulation tasks are set in a tabletop manipulation scene. The workspace consists of a $1.8\,\text{m} \times 1.8\,\text{m}$ table surface at a height of $0.7\,\text{m}$, enclosed within a $10\,\text{m} \times 10\,\text{m}$ room with walls and ceiling that provide realistic visual backgrounds for rendering. The robot is mounted at a fixed base position $(0, 0)$ on the table, and task objects are placed within the robot's reachable workspace. All tasks use the NVIDIA PhysX engine for physics simulation. The physics timestep is $\Delta t = 0.001\,\text{s}$, and the control decimation factor is $15$, yielding an effective control frequency of approximately $66\,\text{Hz}$.

\paragraph{Task details.}\label{supp:tasks}
For each task, we collect $M=300$ teleoperated demonstrations in simulation. At evaluation time, initial object poses are re-sampled from the task randomization ranges below; we ensure evaluation uses distinct random seeds from training so that the initial poses are not reused. Success is determined by task-specific state-based checkers in the simulator.

\textit{Close Box Lid.}
The box is an articulated object with a single revolute hinge joint. The lid is initialized at an open joint angle of $135^\circ$. The box base position is sampled uniformly with $x\in[0.10,0.25]\,\text{m}$ and $y\in[-0.10,0.15]\,\text{m}$. We declare success if the terminal hinge joint angle satisfies $\theta_{\text{lid}} < 14^\circ$.

\textit{Stack Cup into Bowl.}
The bowl mesh is YCB 024\_bowl scaled by $2\times$, and the cup mesh is YCB 065-d\_cups scaled by $2.0\times$. The bowl mass is set sufficiently high that it remains effectively stationary under incidental contacts during insertion. The bowl position is sampled uniformly in a $14\,\text{cm}\times 14\,\text{cm}$ square centered at $(0.15,0.15)\,\text{m}$, and the cup position is sampled uniformly (independently) in a $14\,\text{cm}\times 14\,\text{cm}$ square centered at $(0.15,-0.15)\,\text{m}$; equivalently, $(x_{\text{bowl}},y_{\text{bowl}})\in[0.08,0.22]\times[0.08,0.22]\,\text{m}$ and $(x_{\text{cup}},y_{\text{cup}})\in[0.08,0.22]\times[-0.22,-0.08]\,\text{m}$.

The success checker defines a world-axis-aligned bounding box $\mathcal{B}$ of size $20\,\text{cm}\times 20\,\text{cm}\times 14\,\text{cm}$ anchored to the bowl position with a small positive vertical offset (so that the box lies inside the bowl volume). Let $\{p_j\}$ be a fixed set of predefined keypoints on the cup mesh in world coordinates at the final timestep. We declare success if $p_j\in\mathcal{B}$ for all keypoints. Since $\mathcal{B}$ is axis-aligned and depends only on the bowl position (not its orientation), this criterion is robust to small pose perturbations.

To stabilize cup--bowl contact dynamics, we use task-specific simulation parameters: maximum depenetration velocity $0.5\,\text{m/s}$, contact detection offset $5\,\text{mm}$, and $16$ solver position iterations.

\textit{Kick Football.}
The ball is a sphere with radius $4\,\text{cm}$ and mass $0.1\,\text{kg}$, with static and dynamic friction coefficients $0.1$ and restitution $0.3$. The goal is a rigid mesh asset scaled to approximately $30\,\text{cm}\times 12\,\text{cm}\times 11\,\text{cm}$ (width$\times$height$\times$depth) and assigned a high mass to prevent displacement. The goal position is sampled uniformly in a $14\,\text{cm}\times 14\,\text{cm}$ square centered at $(0.25,0.18)\,\text{m}$ and the ball position is sampled uniformly (independently) in a $14\,\text{cm}\times 14\,\text{cm}$ square centered at $(0.12,-0.10)\,\text{m}$; equivalently, $(x_{\text{goal}},y_{\text{goal}})\in[0.18,0.32]\times[0.11,0.25]\,\text{m}$ and $(x_{\text{ball}},y_{\text{ball}})\in[0.05,0.19]\times[-0.17,-0.03]\,\text{m}$.

The success checker defines a world-axis-aligned bounding box of size $30\,\text{cm}\times 13\,\text{cm}\times 12\,\text{cm}$ placed inside the goal mouth, and declares success if the final ball center lies within the box. We use the same stabilized contact parameters as in ``Stack Cup into Bowl''.

\paragraph{Evaluation protocol.}
We evaluate each simulation method using the task-agnostic procedure in Alg.~\ref{alg:supp-eval}. Concretely, for each of $n_{\text{rollouts}}{=}100$ rollouts we reset the simulator, sample a fresh initial object configuration from the task randomization ranges, and run closed-loop interaction for up to $T{=}500$ control steps by rendering an RGB observation from a fixed evaluation camera, executing actions from the current action chunk, querying the policy for a new chunk when the action cache is exhausted, and stepping the physics. We terminate rollouts early when the task-specific success predicate becomes true, and report the empirical success rate $\hat{P}=s/n_{\text{rollouts}}$.

\begin{algorithm}[t]
\caption{Simulation evaluation (task-agnostic).}
\label{alg:supp-eval}
\begin{algorithmic}[1]
\REQUIRE Simulator environment $\mathcal{E}$; trained policy $\pi_\theta$; evaluation camera $\mathrm{cam}$; task randomizer $\mathrm{SampleInit}(\cdot)$; success predicate $\mathrm{Success}(\cdot)$; rollouts $n_{\text{rollouts}}{=}100$; maximum timestep $T{=}500$
\ENSURE Success rate $\hat{P}\in[0,1]$
\STATE $s \leftarrow 0$ \COMMENT{success counter}
\FOR{$i \leftarrow 1$ \TO $n_{\text{rollouts}}$}
  \STATE $\mathcal{E}.\mathrm{reset}()$
  \STATE $\xi \leftarrow \mathrm{SampleInit}()$
  \STATE $\mathcal{E}.\mathrm{set\_object\_states}(\xi)$ \COMMENT{initialize object poses}
  \STATE $\mathcal{C} \leftarrow \emptyset$ \COMMENT{action cache}
  \FOR{$t \leftarrow 1$ \TO $T$}
    \STATE $o_t \leftarrow \mathcal{E}.\mathrm{render\_rgb}(\mathrm{cam})$
    \IF{$\mathcal{C}=\emptyset$}
      \STATE $\mathcal{C} \leftarrow \pi_\theta(o_{t-k+1:t})$ \COMMENT{predict a chunk of $H$ actions}
    \ENDIF
    \STATE $a_t \leftarrow \mathcal{C}.\mathrm{pop}()$
    \STATE $\mathcal{E}.\mathrm{step}(a_t)$ \COMMENT{advance physics}
    \IF{$\mathrm{Success}(\mathcal{E}.\mathrm{state}())$}
      \STATE $s \leftarrow s + 1$
      \STATE \textbf{break} \COMMENT{terminate rollout early on success}
    \ENDIF
  \ENDFOR
\ENDFOR
\RETURN $\hat{P} \leftarrow s / n_{\text{rollouts}}$
\end{algorithmic}
\end{algorithm}

\subsection{Robots and Control}
\label{supp:robots}

\paragraph{Robot specifications.}
In simulation, we use the Franka Emika Panda as the source robot and evaluate zero-shot cross-embodiment transfer to two target robots with distinct kinematics and appearances: the UFactory xArm7 and the Rethink Robotics Sawyer. In real-world experiments, we instead use xArm as the source and Franka as the target. All three platforms are single-arm manipulators with a 7-DoF revolute arm and a parallel-jaw gripper. For Sawyer, we keep the head-pan joint fixed at its default value ($-2.0943$ rad) throughout all experiments, and only command the 7 arm joints and gripper.

Although the policy operates in a shared end-effector action space (Cartesian pose at the tool center point plus a binary gripper command), these embodiments differ substantially in link geometry, joint limits, gripper mechanics, and visual appearance, creating large embodiment-dependent observation shifts.

\paragraph{Joint control.}
All robots use position-mode joint drives with PD control in simulation. We use stiffness/damping $(80, 4)$ for Franka arm joints and $(800, 40)$ for both xArm7 and Sawyer arm joints. All gripper joints use stiffness $1000$ and damping $100$ to achieve stable opening/closing under contact. All gains are specified in simulator joint-drive units.

\paragraph{TCP offsets and gripper joint values.}
To instantiate the TCP offset matching described in \S~\ref{sec:supp-impl-pose-sampling}, we place the TCP at a fixed translation along the local $+z$ axis of each robot's default end-effector link. This offset is a pure translation with no rotation, and the values below are the offset distances from the named link frame to the TCP.
\begin{itemize}
    \item Franka: 10.3 cm from \texttt{panda\_hand}.

    \item xArm7: 15.0 cm from \texttt{link7}.

    \item Sawyer: 14.1 cm from \texttt{right\_gripper\_base}.
\end{itemize}
The binary gripper command $g\in\{0,1\}$ predicted by the policy corresponds to the gripper dimension of the action vector a = [p, q, g] defined in \S\ref{sec:supp-imp-action-representation}. At inference time, we expand this binary command into robot-specific low-level gripper joint targets. These targets are joint positions rather than end-effector poses; each robot has a different gripper mechanism and a different number of actuated joints, so the mapping is robot-specific. The joint targets for open and closed states for each robot are as follows.
\begin{itemize}
    \item Franka: $[0.0, 0.0]$ for closed, $[0.04, 0.04]$ for open.

    \item xArm7: $[0.85, 0.85, 0.85, 0.85]$ for closed, $[0.0, 0.0, 0.0, 0.0]$ for open.

    \item Sawyer: $[0.0, 0.0]$ for closed, $[0.020833, -0.020833]$ for open.
\end{itemize}

During cross-embodiment transfer, the binary gripper command predicted by the source-trained policy is applied on the target robot using the target robot's open/close mapping above.

\subsection{Data Collection}
\label{supp:data}

\paragraph{Teleoperation system.}
Source demonstrations are collected by a human operator using a teleoperation interface based on DexCap~\citep{wang2024dexcapscalableportablemocap}. The system uses an HTC Vive tracker mounted on the operator's hand to capture arm motion, and a Rokoko smart glove to capture gripper open/close commands. The Vive tracker streams 6-DOF relative poses at $60\,\text{Hz}$. At each control step, the tracker's relative displacement and rotation since a calibration pose are composed with the robot's initial end-effector pose to compute the target end-effector pose $T_\text{target} = T_\text{origin} \circ T_\text{rel}$, where $T_\text{origin}$ is the robot's end-effector pose at the start of the episode and $T_\text{rel}$ is the Vive tracker's relative pose transformed into the robot's coordinate frame via a fixed rotation mapping. An inverse kinematics solver then converts the target end-effector pose to joint commands, which are executed by the simulator. This compositional formulation avoids integration drift that would accumulate with a velocity-based interface. The operator records episodes interactively, with keyboard shortcuts to start recording, stop, save or discard individual episodes, and reset the environment to a new random configuration. Each recorded trajectory stores the initial environment state (object and robot poses), the sequence of joint-space actions, the corresponding end-effector targets (position and quaternion), and optionally the full simulator state at each timestep.

\paragraph{Replay, rendering, and filtering.}
After collection, each trajectory is replayed in the full physics simulator to produce the final observation-action training pairs. During replay, the joint-space actions from the recorded trajectory are executed with physics enabled, and the environment state is advanced by the simulator at each step. The simulator renders RGB observations from a fixed third-person camera in ray-tracing mode, and we record the corresponding end-effector actions (Cartesian poses and gripper commands). Domain randomization is applied once at the beginning of each replay episode, so all frames within a single demonstration share the same randomized appearance. We then validate each replayed trajectory using the task-specific success checker: trajectories that fail to achieve the task criterion
are discarded. Only successfully completed demonstrations are retained for policy training. The resulting filtered dataset of observation--action pairs $\{(o_t^{(\ms)}, a_t^{(\ms)})\}$ serves as training data for the policy. For each simulation task, we collect and retain $M = 300$ successful demonstrations following this pipeline.

\paragraph{Real-world data collection.}
For real-world experiments, demonstrations are collected via teleoperation directly on the source xArm robot at a control frequency of $10\,\text{Hz}$, with episode durations ranging from 7 to 15 seconds. Visual observations are captured using a fixed Intel RealSense D435i camera at a resolution of $640 \times 480$ pixels. The camera extrinsics are kept identical between the source (xArm) and target (Franka) setups to ensure consistent visual input across embodiments. In total, we collect 150 demonstrations for Close Drawer and 61 demonstrations each for Close Box Lid and Wipe Table on the source robot.

\paragraph{Few-shot target demonstrations and real-world evaluation.}
For each real-world task, we collect $10$ target demonstrations on the Franka robot using the same camera viewpoint and control frequency as the source xArm setup. These demonstrations are used for the target-only few-shot baseline and for fine-tuning the source-pretrained policies. During evaluation, each method is tested for $10$ rollouts per task on the Franka robot with randomized initial object configurations. A rollout is counted as successful if the task-specific completion criterion is satisfied within $25$ seconds; otherwise it is recorded as a failure. For \emph{Close Drawer}, success requires the drawer to be fully closed; for \emph{Close Box Lid}, the lid must be closed; and for \emph{Wipe Table}, the target region must be wiped. All methods use the same target demonstrations, evaluation initializations, and camera setup.

\section{Baseline Implementation and Analysis}
\label{sec:supp-baselines}
We first discuss the problem scope in the context of broader cross-embodiment transfer and representation-learning literature, and clarify our problem setting where we isolate the robot-induced visual gap under a shared Cartesian end-effector interface. We then describe two direct visual transfer baselines in the main paper: \textit{Direct Policy Transfer}, which deploys a source-trained visual policy on the target robot without adaptation, and \textit{DINOv2 without observation alignment}, which uses a shared frozen visual encoder for both source and target observations. Next, we detail pixel-space translation baselines that explicitly transform source observations into the target visual domain, including diffusion-based robot-to-robot generation, segmentation-masking, and cross-decoding of aligned features through a pretrained RAE decoder. We further provide qualitative analyses of the translated observations, highlighting artifacts in robot geometry, gripper state, scene context, and contact regions. Finally, we elaborate on the implementation of the real-world RoVi-Aug preprocessing pipeline.

\subsection{Problem Scope and Baseline Rationale}
\label{sec:supp-scope-baselines}

Cross-embodiment transfer aims to reuse policies or demonstrations collected on one robot for another robot with different hardware. Prior work approaches this goal from several directions. One line of work focuses on the action and control interface, for example by conditioning policies on hardware descriptions, defining shared or universal action spaces, training across many embodiments, or adapting demonstrations across morphologies through trajectory optimization~\cite{chen2018hardwareconditioned,yang2024pushinglimits,doshi2024scalingcrossembodied,salhotra2023actionmismatch,zheng2025universalactions}. These approaches reduce kinematic or action-space mismatch, but often rely on strong state/action abstractions, broad multi-robot training data, or task-specific optimization.
In third-person RGB manipulation, however, action alignment alone does not remove the observation-level embodiment gap: even under a shared Cartesian end-effector interface, the same command can produce different robot silhouettes, self-occlusions, gripper appearances, and contact-region evidence that a vision policy must interpret.

A second line of work addresses the visual embodiment gap directly through pixel-space editing or augmentation, such as cross-painting, robot generation, viewpoint augmentation, or segmentation masking~\cite{chen2024mirage,chen2024roviaug,lepert2024shadow}. These methods are closely related to our setting because they explicitly reduce the visual shift induced by changing the robot. However, pixel-space methods require the downstream policy to consume edited or generated observations. As a result, segmentation errors, missing robot pixels, distorted gripper geometry, or inconsistent contact regions can directly affect policy learning and closed-loop execution.

A third line of work learns shared representations from videos or domains, including time-contrastive learning, cross-embodiment inverse reinforcement learning, cross-embodiment skill discovery, video-conditioned imitation, retrieval-based execution alignment, and domain-invariant sim-to-real adaptation~\cite{sermanet2018tcn,zakka2021xirl,xu2023xskillcrossembodimentskill,kedia2025oneshotimitationmismatchedexecution,jain2024vid2robotendtoendvideoconditionedpolicy,tanwani2020dirl}. These methods learn useful invariances across viewpoints, embodiments, or simulation and real domains, but their supervision is usually indirect, relying on temporal contrast, cycle consistency, retrieval, adversarial alignment, or video-level task correspondence. They often require target-domain or multi-domain videos, and the learned representation can be sensitive to data coverage and correspondence assumptions. Their downstream use is also typically narrower than direct robot-to-robot imitation: TCN focuses on viewpoint-invariant representations for imitation or reward learning~\cite{sermanet2018tcn}; XIRL learns a cross-embodiment task-progress embedding used as a dense RL reward~\cite{zakka2021xirl}; and RHyME retrieves and composes unpaired human and robot videos for video-conditioned imitation~\cite{kedia2025oneshotimitationmismatchedexecution}. In contrast, our setting directly reuses source robot observation-action trajectories and learns explicit source-to-target observation alignment from paired robot renderings at matched TCP poses.

\ours focuses on the visual observation gap without synthesizing target-domain RGB demonstrations. After generating paired renderings of source and target robots at the same TCP pose, \ours trains the source encoder to match the embedding produced by a frozen target encoder. This maps each source image to the feature that the target encoder assigns to the corresponding target-robot observation, anchoring source demonstrations in the same latent space used at target deployment. The paired-rendering supervision is also lightweight: it requires robot models, rendering, and inverse kinematics for pose-matched observations, but does not require collecting task-specific target trajectories or training a photorealistic image translator.

We instantiate this problem with tabletop manipulation tasks, 7-DoF arms, parallel-jaw grippers, and a shared Cartesian end-effector action interface. This controlled setting keeps the action interface simple and isolates the observation-level embodiment gap. In the default experiments, the source and target share the same task, camera, and non-robot scene factors; the changed factor is the visible robot embodiment. Thus, the evaluation targets embodiment transfer, not general background or layout robustness: the goal is to preserve source-policy competence when robot body, kinematics, self-occlusions, and contact geometry change. On top of this controlled setting, \ours can still benefit from source-side diversity. Since each source encoder is aligned to the same frozen target encoder, demonstrations from multiple source robots can be aggregated in the shared target-anchored latent space. Similarly, when source demonstrations include randomized backgrounds, the transferred target policy can inherit that visual diversity through the aligned latent representation. 

This scope motivates the baselines used in our experiments. \textit{Direct Policy Transfer} tests whether a source-trained visuomotor policy can bridge the embodiment gap using only the shared action interface. \textit{DINOv2 without observation alignment} tests whether a generic frozen visual representation is already invariant enough to support cross-embodiment policy transfer. Pixel-space translation methods, including \textit{RoVi-Aug}, \textit{Shadow}, and \textit{RoVi-Aug$^\ast$}, address the visual gap by transforming RGB observations instead of aligning representations; we compare their translations qualitatively in \S\ref{sec:supp-translation-qualitative}. \textit{RoVi-Aug$^\ast$} is the strongest controlled pixel-space comparison because it uses the same paired-rendering signal as \ours, but decodes aligned source features back to RGB and trains a pixel-space policy. We therefore use \textit{RoVi-Aug$^\ast$} as the representative pixel-space translation baseline in the simulation-based experiments, where the comparison isolates pixel-space execution from latent-space policy learning. For the real-world few-shot comparison, we separately evaluate the original diffusion-based \textit{RoVi-Aug} pipeline.

\subsection{Direct Visual Transfer Baselines}
\label{sec:supp-direct-visual-baselines}

\paragraph{Direct policy transfer.}
\textit{Direct Policy Transfer} trains a standard source-domain visuomotor policy on source-robot demonstrations and evaluates the same trained policy on the target robot without any adaptation. In simulation, the policy is trained on Franka demonstrations and directly deployed on xArm or Sawyer observations. The policy uses the shared Cartesian end-effector action interface used throughout our experiments, so the baseline isolates the effect of the embodiment-induced visual shift when the action representation and low-level control interface are already shared. The visual encoder and policy are trained end-to-end from source demonstrations using the same action representation and evaluation protocol as the other policies, except that no pretrained DINOv2 encoder, paired-rendering alignment, or target-domain adaptation is used.

\paragraph{DINOv2 without observation alignment.}
This baseline replaces the aligned encoder pair $(E_\ms, E_\mt)$ in \ours~with a single frozen DINOv2 encoder $E_0$ shared by both source and target observations. It does not use paired rendered observations for representation learning, and no source encoder is adapted toward a target-anchored feature space. During policy training, each source demonstration frame is encoded as
\begin{equation}
    z^{(S)}_t = E_0(o^{(S)}_t),
\end{equation}
and the resulting features are pooled and used to train the same flow-matching latent policy as in \ours. At deployment, target observations are encoded by the same frozen encoder,
\begin{equation}
    z^{(T)}_t = E_0(o^{(T)}_t),
\end{equation}
and passed to the trained policy. This baseline tests whether general-purpose DINOv2 features are already sufficiently invariant to robot geometry, appearance, self-occlusion, and gripper-object contact differences. Compared with \ours, the policy architecture, action representation, pooling design, and rollout protocol are kept fixed; the only removed component is target-anchored observation alignment.

\subsection{Pixel-Space Translation Baselines}
\label{sec:supp-baseline-choice}

\begin{figure}[h]
    \centering
    \includegraphics[width=1\linewidth]{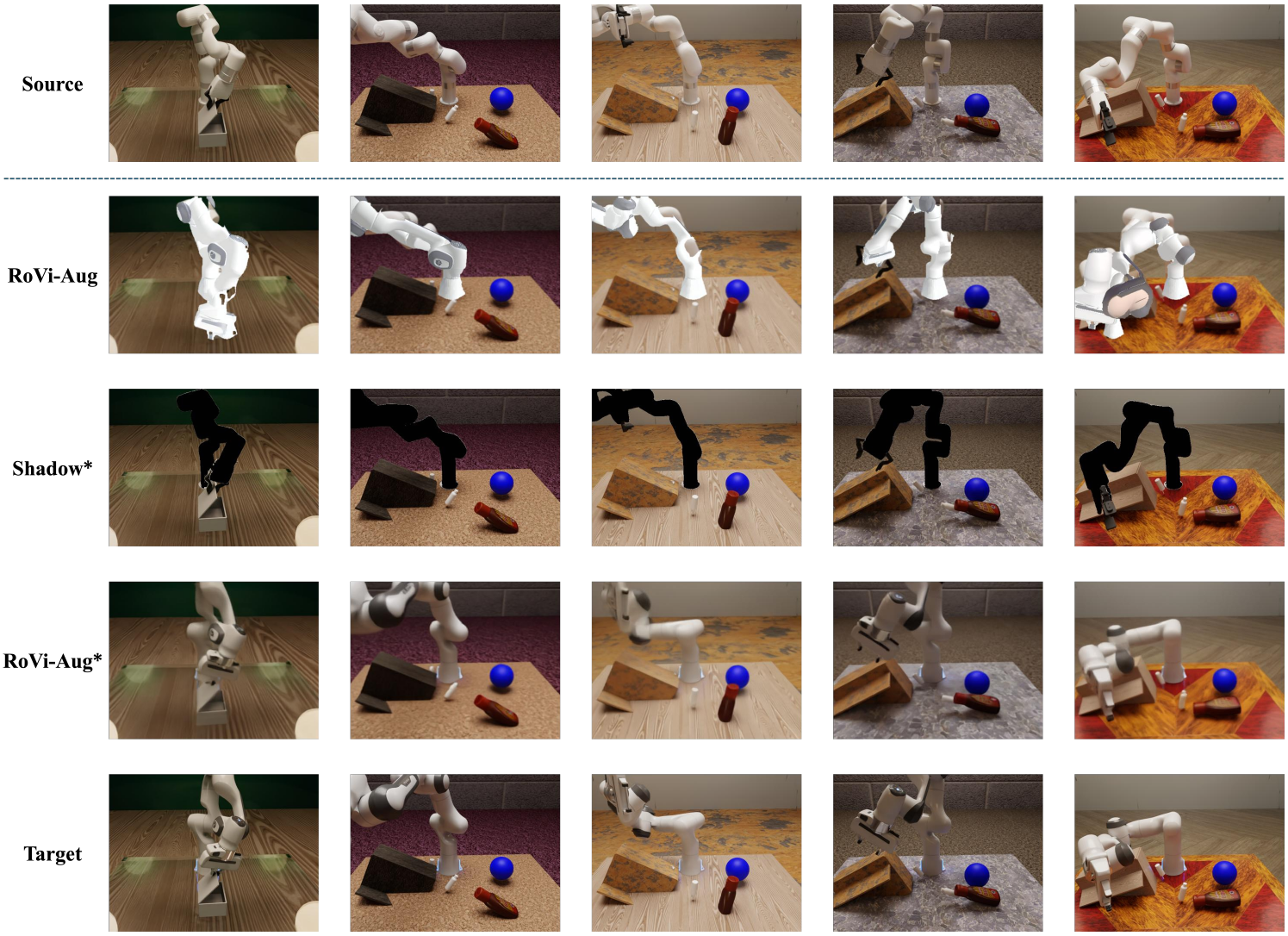}
    \caption{\textbf{Qualitative comparison of pixel-space translation baselines for cross-embodiment transfer.} \textbf{Source} is the input xArm observation; \textbf{Target} is the ground-truth Franka observation under the same end-effector pose. \textbf{RoVi-Aug}, \textbf{Shadow$^\ast$}, and \textbf{RoVi-Aug$^\ast$} show diffusion-based image generation, segmentation-masking, and cross-decoding translations, respectively.}
    \label{suppfig:translations}
\end{figure}

All three translation methods operate on RGB observations, but differ in how they construct target-domain images. \textit{RoVi-Aug} synthesizes the target robot with an image-to-image diffusion model; \textit{Shadow$^\ast$} reimplementation removes the source robot pixels using segmentation masks; and \textit{RoVi-Aug$^\ast$} decodes aligned source features into the target visual domain. We describe each implementation below.

\paragraph{RoVi-Aug: diffusion-based robot-to-robot generation.}
We reproduce RoVi-Aug~\cite{chen2024roviaug} in the xArm (source) $\rightarrow$ Franka (target) setting. \textit{RoVi-Aug} translates source observations into the target embodiment using a conditional image-to-image diffusion model. For robot segmentation, we use Grounded-SAM2~\cite{ravi2024sam2segmentimages,ren2024grounded}. For robot-to-robot generation, we follow RoVi-Aug and train an image-to-image diffusion model that takes an xArm observation as input and outputs a Franka observation under the same end-effector pose. Since the original paper does not release checkpoints or datasets for the xArm-to-Franka setting, we render paired source-target images using PyBullet~\cite{coumans2019} and train the diffusion model on these paired renderings. Concretely, we fine-tune a pretrained Stable Diffusion XL (SDXL)~\cite{podell2024sdxl} with ControlNet~\cite{ControlNet_2023_ICCV} conditioning, and apply the released RoVi-Aug cross-painting pipeline to obtain translated observations $\tilde{o}^{T}$ for downstream pixel-space policy learning.

\paragraph{Shadow$^\ast$: segmentation-masking translation.}
Since Shadow~\cite{lepert2024shadow} is not open source, we re-implement its segmentation-masking procedure as faithfully as possible based on the description in the original paper, and denote our reimplementation as Shadow$^\ast$. We first estimate a segmentation mask for the source robot using Grounded-SAM2 and then remove the robot region to obtain a robot-free observation for policy learning. This method aims to reduce embodiment appearance shift by discarding robot pixels. However, masking can also remove robot-specific visual cues that are important for manipulation, such as gripper pose, self-occlusion, and local contact geometry near the manipulated object.

\paragraph{RoVi-Aug$^\ast$: cross-decoding translation.}
We design RoVi-Aug$^\ast$ as a controlled pixel-space translation baseline that uses the same paired-rendering alignment pipeline as \ours, but executes the downstream policy in RGB space instead of latent space. After learning aligned encoders, we translate a source observation $o^{S}$ by encoding it with the aligned source encoder $E_S$ and decoding the resulting feature with a decoder $D_T$:
\begin{equation}
\tilde{o}^{T} = D_T\big(E_S(o^{S})\big).
\end{equation}
The decoder $D_T$ is the pretrained Representation Autoencoder (RAE)~\cite{zheng2025diffusiontransformersrepresentationautoencoders} decoder, used off the shelf without fine-tuning, which maps DINOv2 latents back to RGB images. Because $E_S$ is trained to reproduce the target encoder's features, decoding $E_S(o^{S})$ yields an image in the target visual domain. We then train a standard pixel-space policy on the translated source demonstrations $\{(\tilde{o}^{T}, a^{S})\}$ using the same end-effector action supervision as the original source demonstrations. At evaluation, the policy directly consumes target-robot RGB observations. Thus, RoVi-Aug$^\ast$ provides a controlled comparison of \emph{pixel-space} and \emph{latent-space} policy learning: it uses the same paired-rendering information as \ours, but trains the policy on translated pixels instead of aligned latent features and deploys it on raw target RGB observations.

\subsection{Qualitative Analysis of Pixel-Space Translation Methods}
\label{sec:supp-translation-qualitative}

Figure~\ref{suppfig:translations} compares the three translation strategies on held-out source-robot observations. Each method takes an xArm observation as input and aims to produce a Franka observation under the same end-effector pose.

The diffusion-based RoVi-Aug baseline often produces visually plausible target-robot images, but the generated robot geometry can be unstable under challenging poses, heavy occlusions, or large kinematic gaps. We observe distorted links, inconsistent joint structure, and occasional gaps between robot parts. These artifacts are problematic for policy learning because the policy is trained directly on the translated RGB observations.

Shadow$^\ast$ avoids hallucinating a target robot by removing source-robot pixels, but this makes it brittle in interaction-heavy frames. In particular, masking can erase visual evidence near the gripper and object, including contact cues that are essential for manipulation. As a result, Shadow$^\ast$ reduces embodiment appearance shift at the cost of discarding task-relevant robot-object interaction information.

RoVi-Aug$^\ast$ produces the most stable translations among the compared pixel-space methods in our setting. Because the source observation is first mapped into a target-aligned latent space before decoding, RoVi-Aug$^\ast$ better preserves scene context and produces fewer structural artifacts around the robot. It also most closely matches the ground-truth target-robot observations among the compared baselines. We therefore adopt RoVi-Aug$^\ast$ as the representative pixel-space translation baseline in our simulation experiments. Since RoVi-Aug$^\ast$ and \ours~use the same paired-rendering alignment signal, their comparison evaluates the practical difference between training a policy on translated RGB observations and training it directly in the aligned latent space.

\subsection{Real-World RoVi-Aug Preprocessing}
\label{sec:supp-real-roviaug-preprocess}

For the \emph{RoVi-Aug pretrain + few-shot} setting in the real-world experiments of the main paper (Sec.~5.2), source xArm demonstration frames are translated into the Franka visual domain before pixel-space policy pretraining. After collecting source demonstrations, we run Grounded-SAM2 on each frame to obtain a source-robot mask. We then apply the RoVi-Aug cross-painting procedure following the original paper: first, the masked source-robot region is removed to form a robot-free canvas; second, the robot-to-robot generation model described in \S\ref{sec:supp-baseline-choice} (SDXL with ControlNet conditioning~\cite{chen2024roviaug,podell2024sdxl,ControlNet_2023_ICCV}) synthesizes the target robot in the same scene context; third, the generated target robot is composited back into the frame to obtain a translated observation $\tilde{o}^{T}$. We use the original diffusion-based RoVi-Aug pipeline for this real-world baseline to compare against the published pixel-space translation method.

This preprocessing is computationally expensive. The robot-to-robot generation step takes approximately 10 seconds per frame. Since our real-world demonstrations are recorded at 10 Hz, translating a 10-second trajectory with 100 frames takes roughly 1000 seconds, or about 17 minutes. This cost is incurred for the source demonstration frames used by the pixel-space translation baseline. In contrast, \ours~does not synthesize target-domain RGB demonstrations: source observations are encoded into the target-aligned latent space for policy learning, and target observations are encoded directly by the frozen target encoder at deployment.

\paragraph{Discussion.}
The few-shot real-world comparison in the main paper includes a diffusion-based \emph{RoVi-Aug pretrain + few-shot} setting. RoVi-Aug pretraining provides a useful transfer signal, improving over training from scratch on the target robot on all three tasks (23/30 versus 11/30 in total). However, \ours~matches or exceeds it on all three tasks, with the largest margin on \textit{Wipe Table}, the most contact-heavy task. One possible explanation is that translation artifacts and inconsistent robot geometry introduce discrepancies between the translated training images and real target observations. \ours~avoids this pixel-space mismatch by training and deploying the policy in the aligned latent space.

\section{Additional Results and Analyses}
\label{sec:supp-additional}

\subsection{Pooling and Feature Design in Latent Policy Training}
\label{sec:supp-align-pooling}

\paragraph{Motivation and scope.}

Our approach uses a pretrained DINOv2 backbone that produces patch tokens $\{h_j\}_{j=1}^{N_p}$ and a global \texttt{[CLS]} token $h_{\mathrm{cls}}$.
In the cross-embodiment setting, we perform observation alignment by optimizing a source encoder $E_S$ to match a frozen target encoder $E_T$ instantiated from pretrained DINOv2 weights (Eq.~(1) in the main paper).
Independently of the alignment objective, a key practical design choice for downstream policy learning is how to pool token-level representations into a compact embedding.
To isolate the effect of pooling, we evaluate different pooling strategies in a single-robot (Franka), simulation-based setting, where no embodiment shift is present.
Throughout this section, ``frozen encoder'' refers to freezing the DINOv2 backbone; any pooling module on top (when used) is treated as part of the policy and trained jointly with the policy network.

\paragraph{Pooling strategies.}
We construct the policy input embedding $\phi(o)$ from DINOv2 tokens using the following pooling operators:
\begin{enumerate}
\item \textbf{Mean}: $\phi(o)=\frac{1}{N_p}\sum_{j=1}^{N_p} h_j$.
\item \textbf{CLS}: $\phi(o)=h_{\mathrm{cls}}$.
\item \textbf{Mean+CLS}: $\phi(o)=\left[h_{\mathrm{cls}};\frac{1}{N_p}\sum_{j=1}^{N_p} h_j\right]$.
\item \textbf{Attn+CLS}: $\phi(o)=\left[h_{\mathrm{cls}};\sum_{j=1}^{N_p}\alpha_j h_j\right]$, where $\alpha_j=\mathrm{softmax}(q^\top h_j)$ and $q$ is a learnable pooling token trained jointly with the policy.
\end{enumerate}

\paragraph{Experimental design.}
We evaluate pooling strategies in simulation on the Franka robot across three tasks, described in \S~\ref{supp:tasks}.
All settings use the same latent policy architecture, training hyperparameters, and evaluation procedure as in the main paper. We vary only the pooling operator used to construct the policy input embedding $\phi(o)$.
We report task success rates over 100 rollouts per task. We consider two training regimes: a \emph{frozen-backbone} setting, and an \emph{end-to-end} setting with an unfrozen backbone.

\paragraph{Results and analyses.}
Table~\ref{tab:supp-pooling-frozen} reports results with a \emph{frozen} DINOv2 backbone, where only the policy is trained (including the learnable pooling token in Attn+CLS).
Attn+CLS substantially outperforms fixed pooling across all tasks, indicating that a lightweight, trainable pooling module is important when the visual backbone is kept fixed.
Based on this analysis, we use Attn+CLS in all experiments in the main paper.

\begin{table}[t]
\centering
\caption{Pooling ablation in a single-robot setting (Franka, simulation) with \emph{frozen} DINOv2 backbone. We report task success rates (\%).}
\label{tab:supp-pooling-frozen}
\setlength{\tabcolsep}{4pt}
\renewcommand{\arraystretch}{1.35}
\begin{tabular}{|l|c|c|c|}
\hline
\textbf{Pooling} & \textbf{Close Box Lid} & \textbf{Stack Cup into Bowl} & \textbf{Kick Football} \\
\hline
Mean      & 10 & 3 & 31 \\
\hline
CLS token & 6  & 0 & 5  \\
\hline
Mean+CLS  & 15 & 3 & 29 \\
\hline
Attn+CLS  & \textbf{97} & \textbf{89} & \textbf{93} \\
\hline
\end{tabular}
\end{table}

We additionally train policies end-to-end with the DINOv2 backbone \emph{unfrozen}.
Although \ours~keeps the DINOv2 backbone frozen during policy training to preserve the target-anchored representation, this experiment examines whether freezing the backbone limits single-robot performance.
As shown in Table~\ref{tab:supp-pooling-unfrozen}, end-to-end training with simple pooling (Mean or Mean+CLS) achieves high success rates.
In comparison, in the frozen-backbone regime, Attn+CLS reaches similarly strong performance among the pooling variants (Table~\ref{tab:supp-pooling-frozen}), suggesting that freezing the backbone can still yield high performance when paired with an appropriate pooling module.
Interestingly, when the backbone is unfrozen, Attn+CLS performs poorly relative to Mean/Mean+CLS, suggesting that higher-capacity pooling can interact unfavorably with end-to-end representation learning and destabilize optimization.
We therefore retain the frozen-backbone design for cross-embodiment transfer and leave end-to-end fine-tuning to future work.

\begin{table}[t]
\centering
\caption{Pooling ablation in a single-robot setting (Franka, simulation) with \emph{unfrozen} DINOv2 backbone trained end-to-end with the policy. We report task success rates (\%).}
\label{tab:supp-pooling-unfrozen}
\setlength{\tabcolsep}{4pt}
\renewcommand{\arraystretch}{1.35}
\begin{tabular}{|l|c|c|c|}
\hline
\textbf{Pooling} & \textbf{Close Box Lid} & \textbf{Stack Cup into Bowl} & \textbf{Kick Football} \\
\hline
Mean      & \textbf{95} & \textbf{92} & 88 \\
\hline
CLS token & 80 & 69 & 79 \\
\hline
Mean+CLS  & \textbf{95} & 91 & \textbf{89} \\
\hline
Attn+CLS  & 15 & 2 & 18 \\
\hline
\end{tabular}
\end{table}

\subsection{Policy Types: Diffusion vs.\ Flow Matching}
\label{sec:supp-policy-types}

\paragraph{Motivation and scope.}
Our policy framework is applicable to different generative policy types operating in the aligned latent space.
The main paper uses a conditional flow-matching policy as the default instantiation.
Here, we implement a DDPM-style diffusion policy~\cite{chi2023diffusionpolicy} and compare it against flow matching under a matched training and evaluation design.
We hold the aligned encoders fixed and vary only the generative formulation, thereby isolating its effect on in-domain performance and cross-embodiment transfer.

\paragraph{DDPM variant.}
The DDPM variant replaces the flow-matching objective with a denoising diffusion objective over action chunks.
Given a demonstrated action chunk $\mathbf{a}$ and Gaussian noise $\boldsymbol{\epsilon}\sim\mathcal{N}(\mathbf{0},\mathbf{I})$, we sample a diffusion step $n\sim\mathrm{Uniform}\{0,\dots,T_{\mathrm{diff}}{-}1\}$ with $T_{\mathrm{diff}}=100$ and construct a noisy action chunk
$\tilde{\mathbf{a}}_n=\sqrt{\bar{\alpha}_n}\,\mathbf{a}+\sqrt{1-\bar{\alpha}_n}\,\boldsymbol{\epsilon}$,
where $\{\bar{\alpha}_n\}$ follows a cosine noise schedule.
The conditional UNet predicts the noise residual $\boldsymbol{\epsilon}_\theta(\tilde{\mathbf{a}}_n,n\mid\mathbf{z}^{(\ms)}_{t-k+1:t})$ and is trained with
\begin{equation*}
\mathcal{L}_{\mathrm{DDPM}}
=\mathbb{E}_{n,\boldsymbol{\epsilon}}\big[\|\boldsymbol{\epsilon}_\theta(\tilde{\mathbf{a}}_n,n\mid\mathbf{z}^{(\ms)}_{t-k+1:t})-\boldsymbol{\epsilon}\|_2^2\big].
\end{equation*}
At inference, action chunks are generated by iterating the reverse diffusion process for $T_{\mathrm{diff}}$ denoising steps (one network forward pass per step).

In contrast, the flow-matching policy integrates the learned ODE for 10 solver steps using a second-order Midpoint method; each solver step evaluates the velocity network twice (at the start and midpoint), resulting in 20 network forward passes in total.
All other components are identical between the two variants: the same conditional UNet backbone, the same frozen aligned encoder with Attn+CLS pooling (\S\ref{sec:supp-align-pooling}), observation history $k=3$, action horizon $H=8$, end-effector action representation, optimizer, batch size, and training schedule.

\begin{table}[h]
\centering
\caption{\textbf{Policy type comparison in the aligned latent space.} Both policies use the same aligned encoders, UNet backbone, training data, and hyperparameters; only the generative formulation differs. Franka is the source robot (in-domain); xArm is the transfer target (Franka$\to$xArm). We report success rates (\%) over 100 rollouts.}
\label{tab:supp-policy}
\renewcommand{\arraystretch}{1.35}
\begin{tabular}{|l|c|c|c|c|c|c|c|}
\hline
 & \multicolumn{2}{c|}{\textbf{Close Box Lid}} & \multicolumn{2}{c|}{\textbf{Stack Cup into Bowl}} & \multicolumn{2}{c|}{\textbf{Kick Football}} & \textbf{Avg.\ } \\
\hline
\textbf{Policy} & \textbf{Franka} & \textbf{xArm} & \textbf{Franka} & \textbf{xArm} & \textbf{Franka} & \textbf{xArm} & \textbf{} \\
\hline
Diffusion (DDPM) & 89 & 85 & 79 & 49 & 84 & 76 & 77.0 \\
\hline
Flow Matching     & \textbf{97} & \textbf{93} & \textbf{89} & \textbf{65} & \textbf{93} & \textbf{89} & \textbf{87.7} \\
\hline
\end{tabular}
\end{table}

\paragraph{Results and analyses.}
Table~\ref{tab:supp-policy} compares diffusion (DDPM) and flow matching on three simulation tasks (\S\ref{supp:tasks}), reporting both in-domain performance on the source robot (Franka) and cross-embodiment transfer to xArm (Franka$\to$xArm), with 100 rollouts per setting.
Flow matching consistently outperforms DDPM across tasks and evaluation settings, improving the overall average success from 77.0\% to 87.7\% (averaged over three tasks and two robots).
Both policy types exhibit consistent task-level trends: success is highest on \textit{Close Box Lid} and \textit{Kick Football}, while \textit{Stack Cup into Bowl} shows the largest transfer gap, reflecting its higher contact sensitivity.
These results indicate that \ours is compatible with different generative policy types in the aligned latent space, while flow matching is a stronger instantiation in our setting.
In addition to higher success, flow matching requires $5{\times}$ fewer network forward passes at inference (20 vs.\ 100 NFEs), yielding substantially faster action generation.
We therefore adopt flow matching as the default policy throughout the paper.

\subsection{Effect of Manipulation-Relevant Pose Sampling}
\label{sec:supp-additional-pose-prior}
We ablate the pose distribution used to generate paired renderings for observation alignment. The proposed sampler draws manipulation-relevant TCP poses using the mixture described in \S\ref{sec:supp-impl-pose-sampling}, combining broad workspace sampling with empirical demonstration-pose samples. As a baseline, we use unconstrained random TCP-pose sampling, which samples feasible TCP poses broadly without biasing them toward task-plausible tabletop manipulation configurations. The downstream policy-training and evaluation protocol is kept fixed. We evaluate zero-shot transfer on \emph{Close Box Lid} from Franka to xArm and Sawyer, using $100$ simulation rollouts per target robot.

\begin{table}[H]
    \centering
    \caption{\textbf{Effect of pose sampling for paired rendering.}
    We train the observation alignment stage with either unconstrained random TCP-pose samples or the proposed manipulation-relevant pose prior, and evaluate zero-shot \emph{Close Box Lid} transfer from Franka to xArm and Sawyer. Entries report success rates (\%) over $100$ simulation rollouts per target robot.}
    \renewcommand{\arraystretch}{1.35}
    \small
    \begin{tabular}{|l|c|c|}
        \hline
        \textbf{Pose generator for paired poses} & \textbf{xArm} & \textbf{Sawyer} \\
        \hline
        Unconstrained random poses & 60 & 32 \\
        \hline
        Manipulation-relevant pose prior & 93 & 81 \\
        \hline
    \end{tabular}

    \label{tab:pose-prior-close-box}
\end{table}

Table~\ref{tab:pose-prior-close-box} shows that manipulation-relevant pose sampling substantially improves transfer performance on both target embodiments. Compared with unconstrained random TCP-pose sampling, the manipulation-relevant prior improves success from $60\%$ to $93\%$ on xArm and from $32\%$ to $81\%$ on Sawyer. For reference, the corresponding in-domain Franka policy reaches $97\%$ success on \emph{Close Box Lid}; thus, manipulation-relevant pose sampling nearly recovers source-level performance on xArm and substantially narrows the transfer gap on Sawyer. These results indicate that paired rendering is most effective when alignment pairs cover robot configurations likely to appear during manipulation.

\begin{figure}[t]
    \centering
    \begin{subfigure}[t]{0.521\linewidth}
        \centering
        \includegraphics[
            width=\linewidth,
            height=0.33\textheight,
            keepaspectratio
        ]{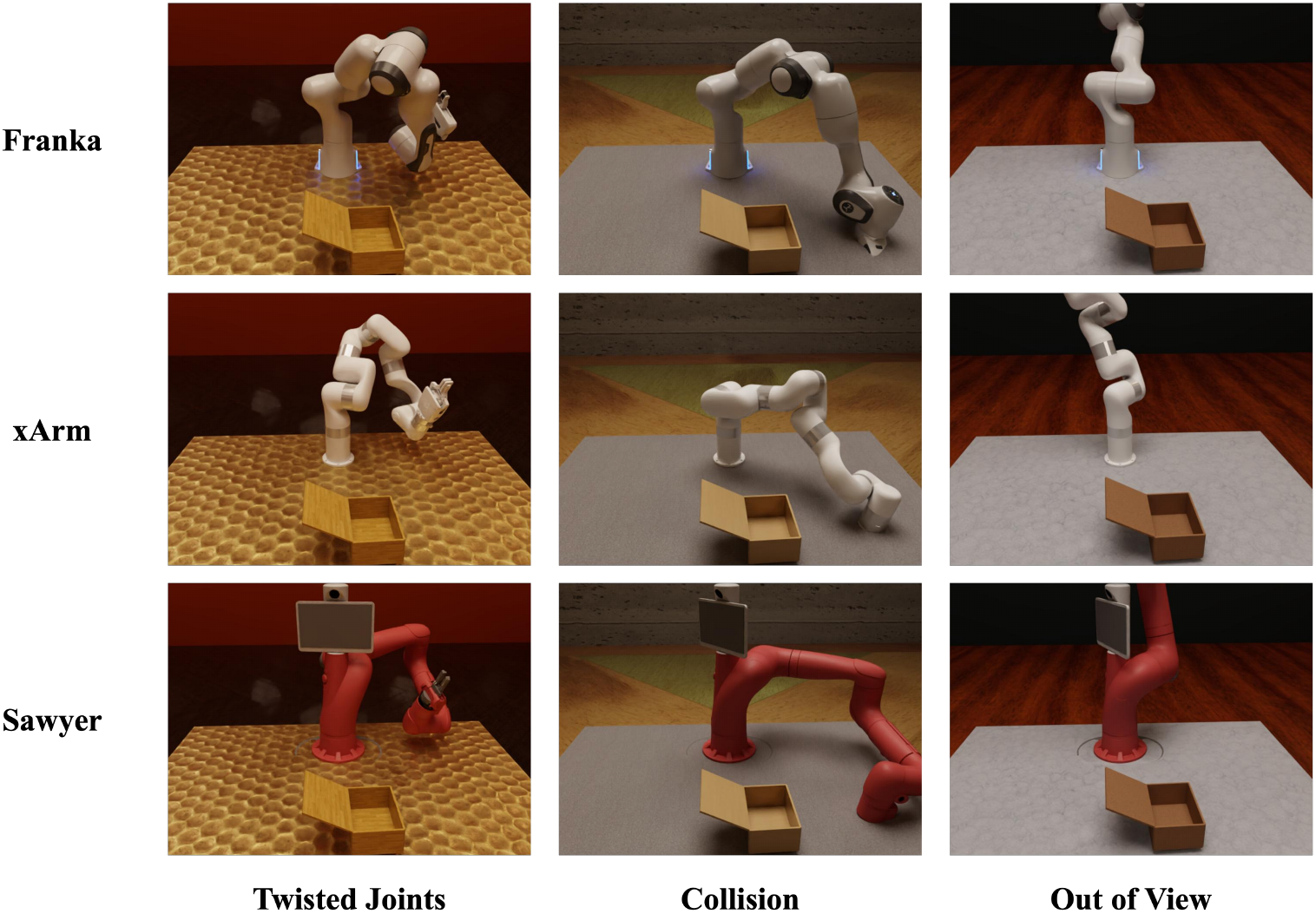}
        \caption{Unconstrained random poses}
        \label{fig:supp-unconstrained-random-poses}
    \end{subfigure}
    \hfill
    \begin{subfigure}[t]{0.455\linewidth}
        \centering
        \includegraphics[
            width=\linewidth,
            height=0.33\textheight,
            keepaspectratio
        ]{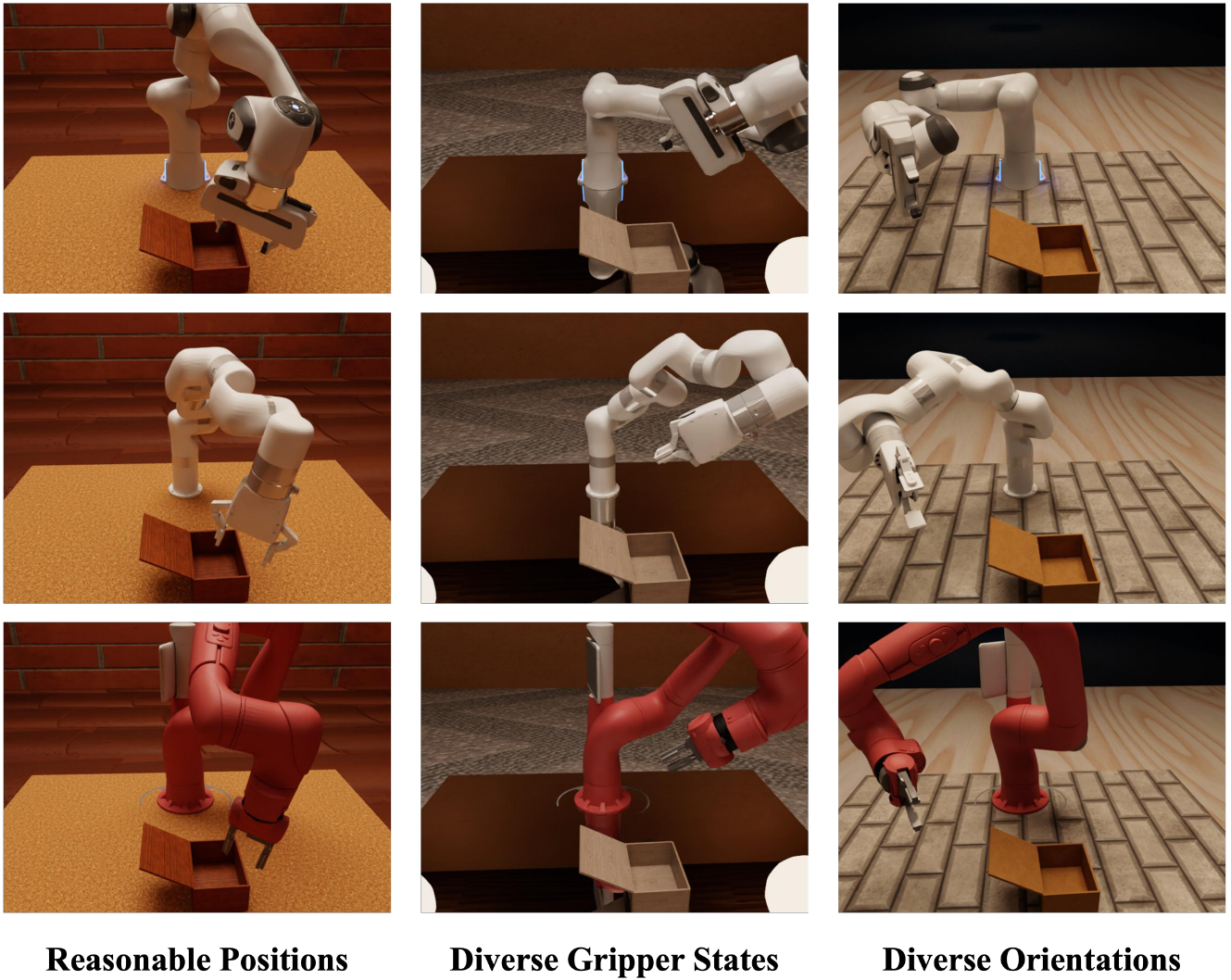}
        \caption{Manipulation-relevant pose prior}
        \label{fig:supp-manipulation-relevant-pose-prior}
    \end{subfigure}

    \caption{\textbf{Qualitative comparison of pose samplers for paired rendering.}
    Unconstrained random pose sampling frequently produces implausible robot configurations, including twisted joints, collisions, and out-of-view grippers. In contrast, the manipulation-relevant pose prior produces task-plausible tabletop configurations with diverse gripper states and orientations, providing more useful paired observations for alignment training.}
    \label{fig:supp-pose-sampling-qual}
\end{figure}

\subsection{Transfer under Simulated Background Shifts}
\label{sec:supp-bg-generalization}

\paragraph{Motivation.}
Cross-embodiment transfer primarily targets the embodiment gap between the source and target robots. At deployment, however, the target robot is often placed in scenes whose background and appearance differ from those seen during source data collection. We therefore examine whether \ours~remains effective when the scene changes between training and deployment, and whether source-side visual diversity can be inherited by the transferred target policy.

\paragraph{Experimental design.}
We study the \textit{Close Box Lid} task with Franka as the source robot and evaluate transfer to xArm and Sawyer. We compare two settings that share the same alignment and latent-policy pipeline and differ only in the scene distribution. In the \emph{matched-background} setting, training and evaluation scenes are drawn from the same background distribution, following the default protocol used elsewhere in the paper. In the \emph{randomized-background} setting, source demonstrations are rendered with randomized backgrounds, and the transferred policy is evaluated in five held-out scenes that are not seen during training; we report the mean success rate across these scenes.

\begin{table}[h]
\centering
\caption{\textbf{Background generalization via image randomization.} Success rates (\%) on the simulated Close Box Lid task, with Franka as the source robot and xArm and Sawyer as transfer targets. The first row trains with randomized source backgrounds and tests in 5 unseen scenes; the second uses matched train--test backgrounds.}

\begin{flushleft}
\renewcommand{\arraystretch}{1.35}
\setlength{\tabcolsep}{3pt}
\centering
\begin{tabular}{|l|c|c|c|}
\hline
 & \textbf{Franka (source)} & \textbf{xArm} & \textbf{Sawyer} \\
 \hline
Randomized background & 88.6 & 85.6 & 80.0 \\
\hline
Matched background & \textbf{97} & \textbf{93} & \textbf{81} \\

\hline
\end{tabular}
\end{flushleft}

\label{tab:supp-close-box-bg}
\end{table}

\paragraph{Results and discussion.}
Table~\ref{tab:supp-close-box-bg} shows that \ours~retains most of its performance under unseen backgrounds. Relative to the matched-background setting, success on the five held-out scenes drops only modestly for the source Franka ($97\% \rightarrow 88.6\%$) and for xArm ($93\% \rightarrow 85.6\%$), and is essentially unchanged for Sawyer ($81\% \rightarrow 80.0\%$). The aligned latent policy therefore retains most of its performance under the tested held-out simulated backgrounds, suggesting that background randomization applied on the source side carries over to the target embodiments through the shared latent space.

More broadly, observation alignment addresses embodiment differences but does not guarantee robustness to unseen visual conditions. These results suggest that source-side visual diversity can support generalization after cross-embodiment transfer, although its effectiveness depends on the coverage of the source training distribution. This is convenient in practice, as the source-side data can be diversified (\eg, via background randomization) and remains compatible with complementary techniques such as domain randomization, co-training, and fine-tuning with additional synthetic or real trajectories.

\end{document}